\documentclass{article} 
\usepackage{iclr2027_conference,times}

\usepackage{amsmath,amsfonts,bm}

\def\eqref#1{equation~\ref{#1}}

\def\1{\bm{1}}

\DeclareMathAlphabet{\mathsfit}{\encodingdefault}{\sfdefault}{m}{sl}
\SetMathAlphabet{\mathsfit}{bold}{\encodingdefault}{\sfdefault}{bx}{n}

\usepackage[hidelinks]{hyperref}
\usepackage{url}
\usepackage{amsmath}
\usepackage{graphicx}
\RequirePackage{fix-cm}
\usepackage{wrapfig}
\usepackage{graphicx}
\usepackage{capt-of}

\usepackage{booktabs}
\usepackage{multirow}
\usepackage[table]{xcolor}

\usepackage{booktabs}
\usepackage{multirow}
\usepackage{array}
\usepackage[table]{xcolor}

\usepackage{booktabs}
\usepackage{multirow}
\usepackage{array}
\usepackage[table]{xcolor}
\usepackage{pgfplots}
\pgfplotsset{compat=1.18}
\usepackage{amsmath}
\usepackage{tikz}
\usepackage{graphicx}
\usepackage{xcolor}

\definecolor{CTBBlue}{HTML}{235FAD}
\definecolor{CTBOrange}{HTML}{CE6B20}
\definecolor{CTBGray}{HTML}{858E99}

\definecolor{CTBBlue}{HTML}{235FAD}
\definecolor{CTBAccent}{HTML}{DDEEFF}
\definecolor{CTBGroup}{HTML}{EAF1F8}
\definecolor{CTBMuted}{HTML}{687A8B}

\newcommand{\ctbTableFont}{%
    \fontsize{8.5}{10.2}\selectfont}

\newcommand{\ctbGroupRow}[2]{%
    \rowcolor{CTBGroup}
    \multicolumn{#1}{@{}l}{%
        \textcolor{CTBBlue}{\bfseries #2}}\\}

\newcommand{\ctbModelCite}[1]{%
    \hspace{2pt}%
    {\fontsize{6}{7}\selectfont
    \textcolor{CTBMuted}{\citep{#1}}}%
}

\title{TextCanvasBench: A Comprehensive Benchmark for Complex and Long-Text Image Generation
}

\author{\small%
\mbox{%
\href{https://openreview.net/profile?id=~Zhaoqianqian1}{Zhaoqianqian}\textsuperscript{1,*,$\S$},
\href{https://openreview.net/profile?id=~Fan_Li26}{Fan Li}\textsuperscript{1,2,*,$\ddagger$},
\href{https://openreview.net/profile?id=~Yuanlin_Zhang5}{Yuanlin Zhang}\textsuperscript{1},
\href{https://openreview.net/profile?id=~Yuan_Wang44}{Yuan Wang}\textsuperscript{1},
\href{https://openreview.net/profile?id=~Renjing_Pei1}{Renjing Pei}\textsuperscript{2},
\href{https://openreview.net/profile?id=~Jieyu_Yuan1}{Jieyu Yuan}\textsuperscript{1},
\href{https://openreview.net/profile?id=~Chun-Le_Guo1}{Chun-Le Guo}\textsuperscript{1,$\dagger$}}\\[0.4em]
\normalfont\textsuperscript{1} Nankai University (nankai.edu.cn)\\
\normalfont\textsuperscript{2} Huawei Noah's Ark Lab
}

\iclrfinalcopy

\usepackage{xcolor}
\usepackage{pdflscape}
\usepackage{tabularx}
\usepackage{colortbl}
\usepackage{CJKutf8}
\usepackage{wrapfig}
\usepackage{capt-of}
\usepackage{placeins}
\usepackage{graphicx}

\usepackage{placeins}

\begin{document}
\maketitle
\begingroup
\renewcommand{\thefootnote}{}
\makeatletter
\long\def\@makefntext#1{\noindent#1}
\makeatother
\footnotetext{%
\textsuperscript{*}Equal contribution.\quad
\textsuperscript{$\dagger$}Corresponding authors.\quad
\textsuperscript{$\ddagger$}Project lead.\\
\textsuperscript{$\S$}Work done during an internship at Huawei Noah's Ark Lab.}
\endgroup
\fancyhead{}
\renewcommand{\headrulewidth}{0pt}
\raggedbottom

\begin{abstract}
Recent open- and closed-source text-to-image (T2I) models can generate text-rich images, including scientific figures and presentation slides, yet benchmark evaluation has not kept pace with this capability.
Existing text-rendering benchmarks typically focus on short text, a small number of text regions, and image-level or OCR-based aggregate metrics, limiting their ability to assess long-form content, multi-region layouts, and block-level stylistic and semantic relations in practical T2I scenarios.
To address these gaps, we introduce \textsc{TextCanvasBench}, a benchmark of 1,600 specification-guided samples spanning $8$ domains and $38$ subcategories of real-world text-rich visuals, from commercial posters and academic slides to digital interface panels.
The required text ranges from short labels to more than 1,000 characters, and the annotations identify text blocks and their roles.
We evaluate generation through four complementary dimensions, moving from the whole image to individual text blocks and regions.
At the macro image level, we assess glyph quality, the completeness and accuracy of the required text, and whether the image conveys the prompt’s overall meaning. At the micro block and region level, we assess position, orientation, arrangement and reading order, font, size, and color, as well as selected facts and associations tied to individual blocks.
In our experiments, we evaluate nine state-of-the-art open- and closed-source T2I models (e.g., Boogu-Image and GPT-Image-2). The results show that \textsc{TextCanvasBench} covers a wider range of practical text-rich scenarios and reveals differences in long-text rendering capabilities that existing benchmark scores can obscure. Its detailed scores distinguish models with similar scores on other benchmarks, while its rankings agree more closely with human judgments than those of most benchmarks in our comparison. Code will be released at \url{https://github.com/nqian1/TEXTCANVASBENCH}.

\end{abstract}

\section{Introduction}
\label{sec:introduction}
\begin{figure*}[t]
    \centering
    \includegraphics[width=\textwidth]{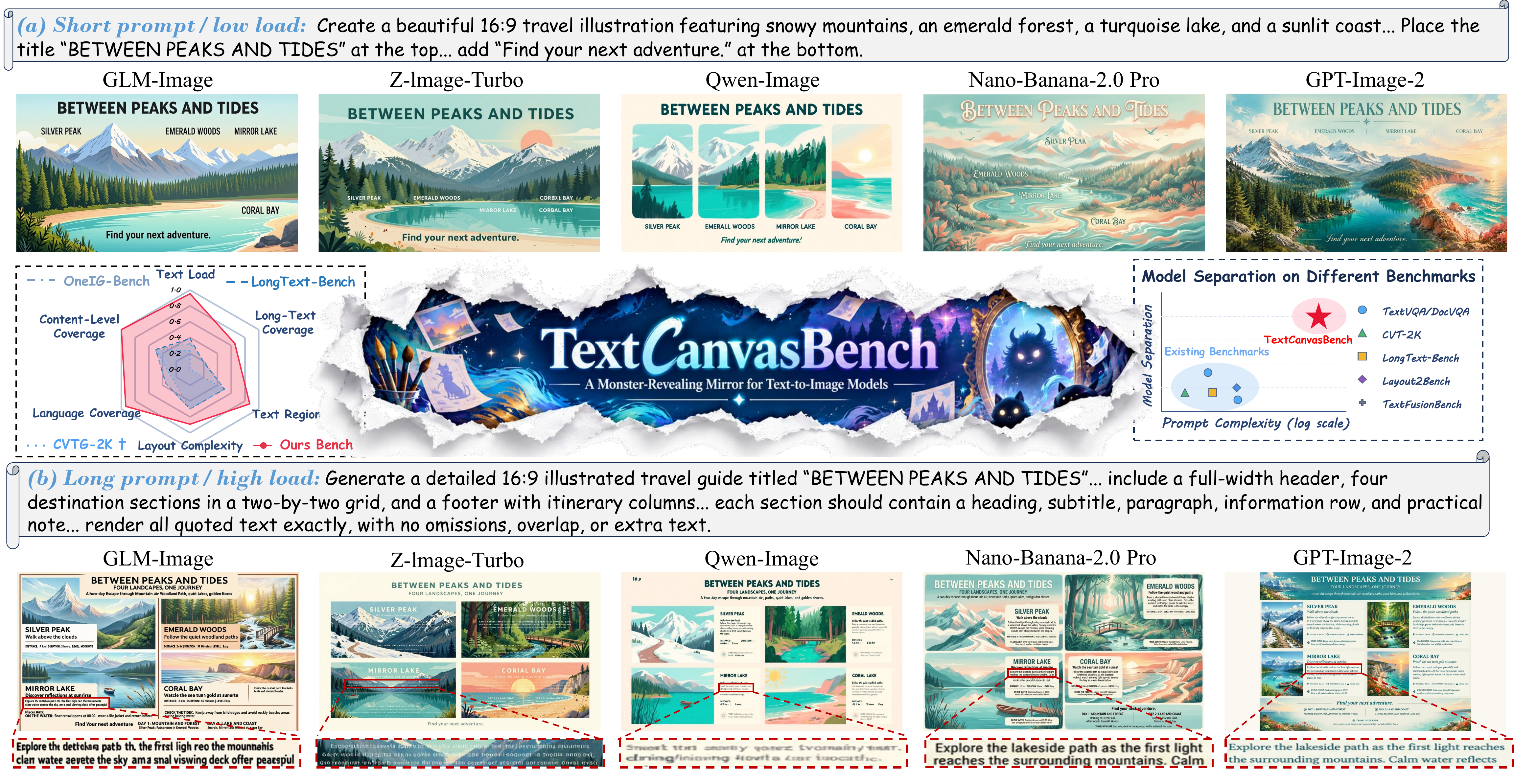}
    \caption{\textbf{Short prompts can hide substantial model differences.}
    (a) With a short prompt and few required text elements, open- and
    closed-source models produce similarly coherent travel illustrations.
    (b) When the same task specifies more text, multiple content sections, and
    a detailed layout, omissions, corruptions, and organization failures become
    more visible, particularly in the open-source outputs. The middle panels
    compare benchmark coverage across six aspects (left) and model separation
    on different benchmarks (right).}
    \label{fig:teaser}
\end{figure*}

Recently, text-to-image (T2I) models~\citep{qwen2025image2512, openai2026gptimage2} can now generate text-rich visuals, from scientific figures and presentation slides to posters and digital interfaces. To be useful, these images must render extended passages, numbers, and units accurately, organize multiple text elements, and link them to the right visual content. 
Existing benchmarks assess text rendering, layout, and prompt following~\citep{wang2025textatlas5m,mao2026textground4m,li2026bizgeneval}. Yet practical text-rich images may require hundreds or thousands of characters across many text blocks, whereas the English portion of LongText-Bench predominantly requires a few dozen words~\citep{geng2025xomni}. Aggregate scores can also conceal differences in text completeness, presentation, and semantic fidelity between models with similar benchmark scores, as illustrated in Figure~\ref{fig:teaser} and quantified in Table~\ref{tab:exp-cross-benchmark}. This motivates an evaluation that tests content, organization, and meaning together while showing where each model fails.

As shown in Figure~\ref{fig:teaser}, all five models produce visually coherent travel illustrations from the short prompt. When the same theme becomes a detailed travel guide with four destination sections, paragraphs, information rows, and a structured footer, their outputs diverge: the open-source models exhibit more omissions, corrupted text, and layout problems, whereas the closed-source models preserve more of the requested content and organization. This contrast exposes an evaluation gap: short prompts can make models appear similarly capable, highlighting the need for a new benchmark that covers a broader mix of text lengths, long-form, text regions, layout complexity, languages, and content requirements, and more clearly differentiates the evaluated models than existing benchmarks.

We therefore introduce \textbf{\textsc{TextCanvasBench}}, a benchmark of 1,600 specification-guided samples spanning $8$ domains and $38$ subcategories of real-world text-rich visuals, from commercial posters and academic slides to digital interface panels. 
Each sample includes a self-contained generation prompt, the complete required text, and reviewed annotations identifying individual text blocks and their roles.
The required text ranges from short labels to over 1,000 characters. Models generate images using only the prompt and required text while reference images and annotations are used to build and evaluate the benchmark.

Additionally, we evaluate generation through four complementary dimensions, moving from the whole image to individual text blocks and regions.
At the macro image level,
\textbf{D1} measures glyph structure quality, while \textbf{D2} measures whether the complete required text is rendered accurately. At the micro block
level, \textbf{D3} evaluates prompt-specified position, block orientation, arrangement and reading order, font size, style and color.
\textbf{D4} evaluates whether the rendered text preserves the required facts,
intent, text--image associations, scene logic, and integration with specified
text-bearing objects. D3 and D4 use fixed specifications derived before model
comparison and observations collected from each generated image. 
This two-level evaluation distinguishes errors in the rendered text, errors in its layout and style, and errors in the meaning it conveys in context.

We evaluate nine open- and closed-source T2I models, including Boogu-Image and GPT-Image-2. \textsc{TextCanvasBench} covers a wider range of practical text-rich scenarios and distinguishes models with similar scores on other benchmarks. The resulting model rankings agree more closely with human judgments than those of most benchmarks in our comparison, while the four diagnostic dimensions identify errors in glyphs, text content, presentation, and meaning.

Our main contributions are:
\begin{itemize}
    \item \textbf{\textsc{TextCanvasBench}.}
    We introduce 1,600 specification-guided samples across $8$ domains and $38$ subcategories. Each sample includes a generation prompt, required text, reviewed annotations of text blocks and their roles, and fixed evaluation specifications defining the required text presentation and semantic relationships.

    \item \textbf{Macro and micro-level Evaluation.}
    We define four complementary dimensions: D1 for glyph quality, D2 for complete-text accuracy, D3 for layout and style, and D4 for semantic alignment from the whole image to text blocks and regions.

    \item \textbf{Benchmark Validation.}
    Evaluation across nine generators shows that TextCanvasBench distinguishes models with similar performance on existing benchmarks and yields rankings consistent with human judgments. Further analysis demonstrates that it distinguishes textual correctness, adherence to presentation requirements, and semantic fidelity, detecting presentation violations and incorrect semantic associations even when the text itself is correct.
\end{itemize}

\section{TextCanvasBench Construction}
\label{sec:cb-construction}

This section presents the three-stage construction pipeline of
\textsc{TextCanvasBench}. First, we curate the reference images and characterize
the resulting task distribution (Section~\ref{sec:cb-sources}). Next, we convert
each reference image into paired Chinese and English generation prompts that
specify the required text, visual content, and organization
(Section~\ref{sec:cb-annotation}). Finally, we construct frozen D3 and D4
specifications from the finalized prompts, complete required text, and reviewed
annotations. D3 formalizes presentation requirements, whereas D4 defines
semantic conditions and the visual evidence needed to verify them. (Section~\ref{sec:d3_d4_spec})

\subsection{Data Collection and Statistics}
\label{sec:cb-sources}
\label{sec:cb-scenes}
\label{sec:cb-statistics}

\begin{figure}[!tb]
    \centering
    \includegraphics[
        width=\linewidth,
        height=\textheight,
        keepaspectratio
    ]{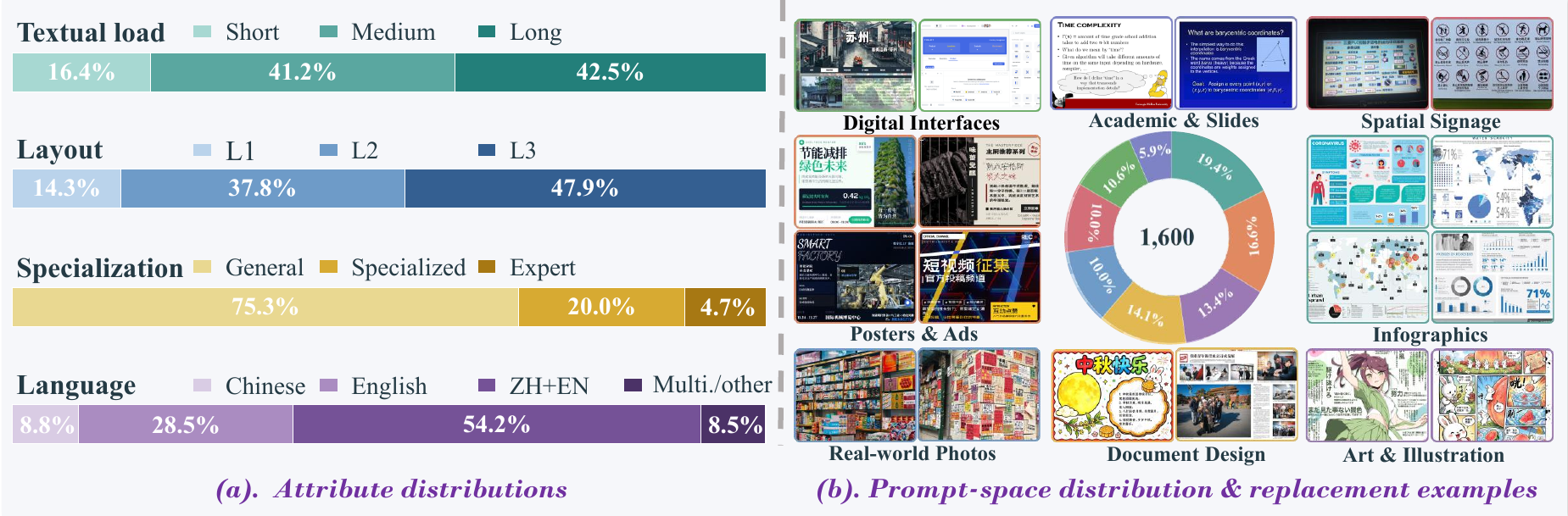}
    \caption{\textbf{Composition of TextCanvasBench.} (a) Distributions of
    required-text length, layout level, content specialization, and language.
    (b) Domain proportions and representative examples across the eight
    application domains.}
    \label{fig:cb-dataset-overview}
\end{figure}

We assembled an initial pool of approximately 10,000 candidate reference
images through image search over publicly accessible websites. We retained
images in which the text is legible and conveys substantive information,
while preserving variation in both content and visual presentation. This
curation process yielded 1,600 reference images.
To represent distinct forms of textual organization in practical images, we
grouped the curated set by visual medium and application context into eight
domains. \emph{Infographics}
combine explanatory passages, numerical labels, legends, and charts across
aligned regions. \emph{Posters \& Ads} use strong typographic hierarchies and
free-form compositions to separate headlines from supporting copy.
\emph{Art \& Illustration} embeds stylized text among visual elements, often
with irregular placement, scale, or orientation. \emph{Document Design}
organizes paragraphs, columns, and tables under explicit reading orders.
\emph{Real-world Photos} place text on physical surfaces under variations in
perspective, illumination, occlusion, and scale. \emph{Spatial Signage}
distributes short functional text across signs whose meaning depends on its
spatial context. \emph{Digital Interfaces} pack small labels, values, and
controls into repeated grids or cards with precise local associations.
\emph{Academic \& Slides} interleave technical text, equations, figures,
captions, and annotations in dense multi-region layouts.
Within each domain, content topic and use case further define 38
subcategories. Individual subcategories contain 24--84 samples, extending
each visual medium across multiple concrete applications rather than a single
content type. Appendix~\ref{app:cb-category-details} provides the complete
taxonomy and sample counts. The resulting collection contains 43,390 required-text instances organized into 12,399 functional regions across eight domains and exhibits broad variation in required-text length, layout complexity, content specialization, and language; detailed dataset statistics
are reported in Figure~\ref{fig:cb-dataset-overview} and Appendix~\ref{app:cb-data-statistics}.

\subsection{Generation Prompts}
\label{sec:cb-annotation}
\label{sec:cb-attribute-rubric}
\label{sec:cb-quality}

We convert each reference image into paired Chinese and English generation
inputs, as illustrated in Figure~\ref{fig:cb-construction-overview}(c). Gemini 2.5 Pro supplies
initial scene descriptions and text transcriptions, which are revised against
the reference image to align the descriptions, required wording, and text
regions. Each input contains a scene description, a region-organized text
inventory, and rendering instructions. The description specifies the visual
content, composition, and selected presentation requirements; the inventory
preserves the required strings and their associations within titles,
paragraphs, tables, and other functional regions.

Chinese and English descriptions express the same task requirements and share
the original-language text inventory. Changing the description language does
not translate the displayed text. For tables, row boundaries and column order
specify which values belong to each subject and attribute. Region identifiers
and formatting delimiters organize the input rather than appearing as text in
the output. Generators receive these textual inputs without the reference
image. Appendix~\ref{app:cb-rubrics} describes the input structure.

\subsection{Evaluation Specification Construction}
\label{sec:d3_d4_spec}

We construct sample-specific D3 and D4 checklists from the finalized prompts
and required text. OCR- and VLM-assisted reference annotations provide text
identities and region membership, distinguishing repeated occurrences and
associating related items. These annotations assist target identification;
reference-image properties absent from the final input do not become evaluation
requirements. Construction records retain source evidence and revisions, and
the released checklists are fixed across evaluated generators.

\textbf{D3 specification.}
Explicit text-presentation requirements are organized into six attribute
families: placement, spatial arrangement and reading order, size hierarchy,
orientation, foreground color, and font style. Each check specifies the
requirement, its target instances or regions, the relevant quantifier, and a
criterion. Quantifiers distinguish requirements on every instance, selected
members, and whole regions. Style requirements retain their specified scope within each region. Source review identifies unsupported requirements and
incorrect target associations before export. The released D3 specifications
contain 33,262 checks over 1,600 samples, as detailed in Appendix~\ref{app:cb-prompt-spec}.

\textbf{D4 specification.}
D4 organizes checks into core textual semantics, communicative intent,
text--image consistency, semantic logic, and physical-carrier integration.
These are five categories, not five fixed questions. Construction aims to
cover the substantive information in the complete required text and the
prompt-supported semantic relations, with no fixed fact-count limit.
Independent propositions are separated, while subjects, attributes, values,
units, conditions, negations, and referents remain bound within each fact.
For tables, a check binds a row subject to its column attribute and value;
for passages, it retains the meaning of a complete assertion rather than
checking isolated words. Equivalent repetitions are merged, and exclusions
are recorded with reasons. Semantic equivalence is allowed without relaxing
required identities, values, or relations.

Each D4 check records its input source, target, satisfaction conditions, and
the generated-image evidence needed for assessment. Physical-carrier checks
apply to specified text-bearing objects and assess attachment and integration
with the depicted surface, without importing unrequested reference details.
The release contains 39,387 checks, including 28,914 core-textual-semantic
checks. Appendix~\ref{app:cb-d4-spec} details construction, and
Appendix~\ref{app:d4-coverage-extension} quantifies necessary-fact coverage,
and Table~\ref{tab:d4-error-detection} reports detection on controlled image pairs.
Actual observations and scores are obtained separately from generated images.

\section{TextCanvasBench Evaluation}
\label{sec:evaluation}

\label{sec:cb-dimensions}
\label{sec:instance-correspondence}

\begin{figure}[!tb]
    \centering
    \includegraphics[
        width=\linewidth,
        height=\textheight,
        keepaspectratio
    ]{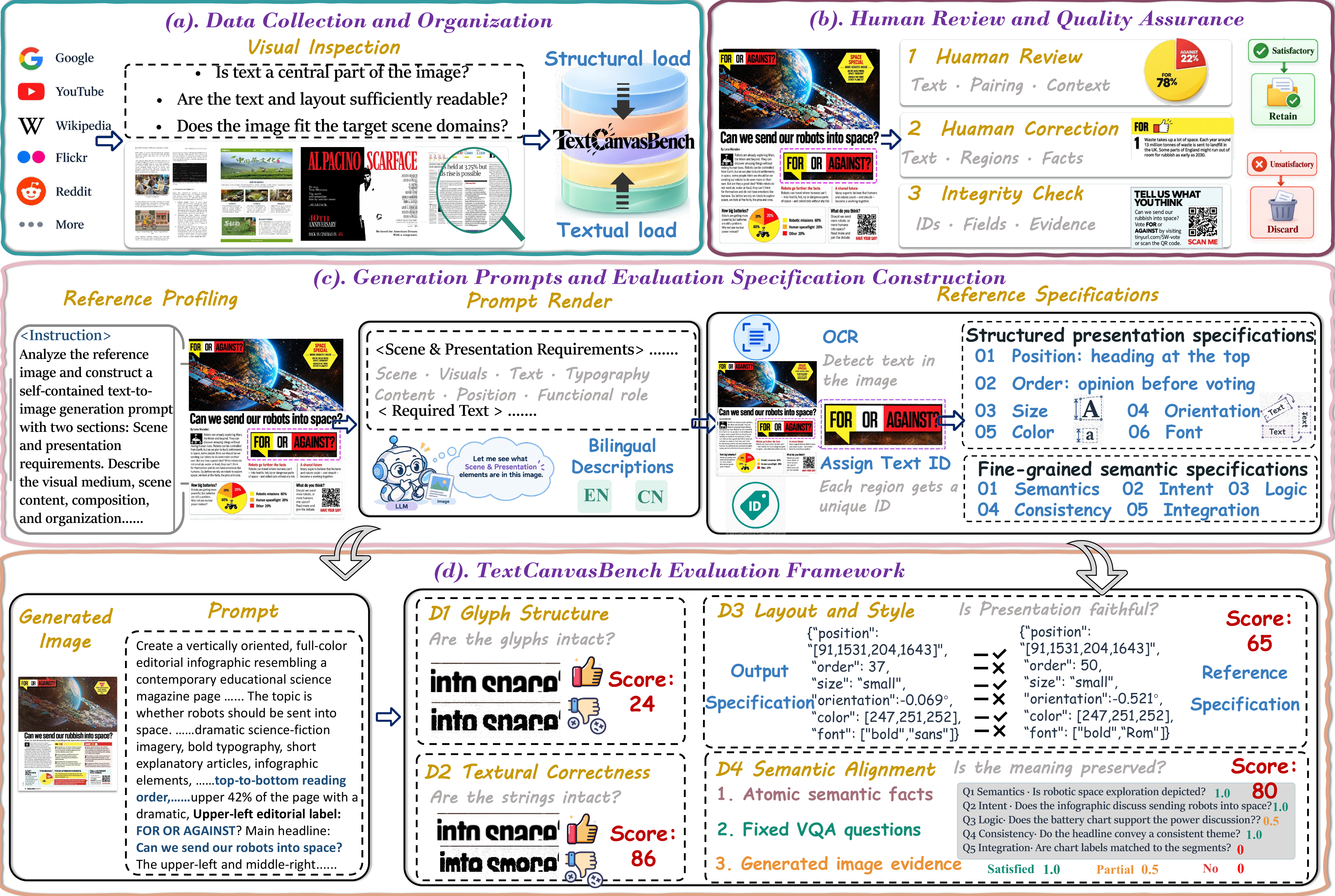}
    \caption{\textbf{Overview of data collection, benchmark construction,
    and evaluation.}
    Collected reference images are screened and annotated to produce
    generation inputs and evaluation specifications. Model outputs are
    then assessed along four diagnostic dimensions using evidence from
    the generated images (Section~\ref{sec:evaluation}).}
    \label{fig:cb-construction-overview}
\end{figure}

Text accuracy alone does not capture failures in visual organization or
semantic association. Therefore, we evaluate generation at two complementary
levels using four diagnostic dimensions as shown in Figure~\ref{fig:cb-construction-overview}(d). 
We use the previously frozen D3 and D4 specifications to determine what must be
verified for each sample (Section~\ref{sec:d3_d4_spec}). During
evaluation, D1 and D2 directly analyze the generated image and its text
transcription. For D3 and D4, a VLM inspects the generated image to obtain the
presentation and semantic evidence required by the corresponding
specifications. OCR candidates can assist text localization and target
association. These observations are then
compared with the specified requirements to obtain the corresponding scores.
Appendix~\ref{app:eval-scoring} provides the detailed evaluation rules.

\subsection{Image-Level Text Evaluation: D1 and D2}
\label{sec:d1}
\label{sec:d2}

For each sample $i$, TextPecker~\citep{zhu2026TextPecker} receives the full
generated image $I_i$ and transcribes its visible text with glyph-error markers.
The complete required text $T_i$ from the final generation prompt is used as the
content reference, retaining repeated occurrences. Let $\widehat T_i$ denote
the parsed transcription. We use TextPecker's glyph-quality function $Q$ and
matching-based normalized edit distance $\operatorname{GNED}$ to obtain
\begin{equation}
 D1_i=100\,Q(\widehat T_i),\qquad
 D2_i=100\bigl[1-\operatorname{GNED}(\widehat T_i,T_i)\bigr].
 \label{eq:eval-text}
\end{equation}
D1 measures glyph integrity through the proportion of marked structural errors.
D2 measures textual correctness through edit costs and unmatched-text penalties,
accounting for substitutions, missing text, and recognized extra or repeated
text across the image. Thus well-formed but incorrect text can receive high D1
and low D2. Both scores are computed directly at macro image level; micro block is used for presentation and association analysis in D3/D4.

\subsection{Layout and Style Evaluation: D3}
\label{sec:d3}

D3 asks whether the generated text follows the layout and style specified in
the prompt. Its six attribute families are placement, spatial/reading order,
size hierarchy, orientation, foreground color, and font style, as detailed in Appendix~\ref{app:cb-prompt-spec}. Frozen checks target individual text
instances, specified groups, or functional regions. Each check receives a
\texttt{pass}, \texttt{fail}, or \texttt{unknown} judgment against its criterion
and target scope. Supported numerical placement constraints are checked
against verified normalized geometry; other requirements are assessed from
the visible text and its context. Unspecified appearance is unpenalized.

\textbf{Generated-image evidence.}
The VLM associates visible text with the specified targets using text identity,
region membership, and surrounding context, with OCR candidates when available.
Each judgment records its supporting observations and target boxes.
Confirmed omissions fail the corresponding requirement; ambiguous target
associations and insufficient evidence remain unknown. Independent assessments
verify the judgments, with a third assessment resolving disagreements when
two evidence-valid votes agree. Unresolved checks receive bounded reinspection
using full-image context and local detail views when available. Technical
failures are retried separately from visual uncertainty, as detailed in Appendix~\ref{app:eval-d3-rules}.

Let $\mathcal C_{ik}$ denote the checks in attribute family $k$ and
$\mathcal A_i$ the families with at least one check. Each resolved check has
$s_{ic}=1$ if satisfied and $s_{ic}=0$ otherwise. When all applicable checks
are resolved, the attribute score $a_{ik}$ and D3 score are
\begin{equation}
 a_{ik}=\frac{1}{|\mathcal C_{ik}|}\sum_{c\in\mathcal C_{ik}}s_{ic},
 \qquad
 D3_i=\frac{100}{|\mathcal A_i|}\sum_{k\in\mathcal A_i}a_{ik}.
 \label{eq:eval-d3}
\end{equation}
The attribute score is the proportion of satisfied requirements: $1$, $0.5$,
and $0$ indicate that all, half, or none of its checks are satisfied,
respectively. These values illustrate the scale rather than restrict it to
three levels. Active families receive equal weight, so additional checks
within one family do not increase its weight relative to the others.

\subsection{Semantic Evaluation: D4}
\label{sec:d4}

D4 spans both evaluation levels: it tests whether the generated image conveys
the intended meaning and whether individual text blocks preserve the required
facts and relationships to the scene. It uses five components: core textual semantics,
communicative intent, text--image consistency, semantic logic, and
physical-carrier integration. Its frozen specification decomposes text-related
meaning into atomic semantic facts that preserve entities, attributes, values,
units, negations, conditions, and associations, together with additional
prompt-supported semantic criteria, as detailed in Appendix~\ref{app:cb-d4-spec}.
Physical-carrier integration applies only to prompt-specified physical
text-bearing surfaces; font or position changes constitute semantic errors only when they
alter required meanings or associations.

\textbf{Generated-image evidence.}
For each check, the evaluator records the actual text read, visible scene
content, and the inspected scope. Association checks additionally record the
participating entities and their observed relation. A judgment is accepted
only when its evidence supports a \texttt{pass} or \texttt{fail} verdict;
otherwise it remains \texttt{unknown}. Confirmed omissions, incorrect facts,
and generation illegibility that prevents recovery of the required meaning
receive failing judgments. Physical-carrier judgments instead depend on the visible relation between
text and its supporting surface. Independent verification and targeted
reinspection address conflicting or incomplete observations, as detailed in Appendix~\ref{app:eval-other-scores}.

Let $\mathcal F_{ik}$ denote the checks in semantic component $k$.
A resolved check has $q_{ic}=1$ if satisfied and $q_{ic}=0$ otherwise.
When all applicable checks are resolved, the component score is
\begin{equation}
 r_{ik}=\frac{1}{|\mathcal F_{ik}|}\sum_{c\in\mathcal F_{ik}}q_{ic}.
 \label{eq:eval-d4-component}
\end{equation}
As in D3, $r_{ik}$ measures the proportion of satisfied requirements; an
intermediate value reflects satisfaction of some requirements within the
component. With base weights $v=(0.45,0.10,0.20,0.20,0.05)$ and applicable
components $\mathcal B_i$, the score is
\begin{equation}
 D4_i=100\sum_{k\in\mathcal B_i}\widetilde v_{ik}r_{ik},\qquad
 \widetilde v_{ik}=\frac{v_k}{\sum_{\ell\in\mathcal B_i}v_\ell}.
 \label{eq:eval-d4}
\end{equation}
Components without checks in the frozen specification are excluded and the
remaining weights are renormalized. Unresolved observations remain unknown;
they neither receive an intermediate score nor make a component inapplicable.

\subsection{Overall Score and Reporting}
\label{sec:aggregate-metric}

All dimensions range from $0$ to $100$. Samples with four valid, applicable
scores receive the equally weighted summary
\begin{equation}
 S_i=0.25D1_i+0.25D2_i+0.25D3_i+0.25D4_i.
 \label{eq:eval-overall}
\end{equation}
We report the four dimension scores alongside $S_i$ so that the summary does
not hide different failure modes. Missing or N/A dimensions are neither
zero-filled nor reweighted to produce $S_i$.

For D3 and D4, a complete point score requires all applicable checks to be
resolved. Remaining unknowns stay in the denominator. For an attribute or
component containing $P$ passing, $F$ failing, and $U$ unknown checks,
check coverage is $(P+F)/(P+F+U)$, and its satisfaction proportion lies between
$P/(P+F+U)$ and $(P+U)/(P+F+U)$. Image-level bounds use the same aggregation
as the complete score. Model means over complete-score images are reported
separately from bounds over all applicable evaluated images. Appendix~\ref{app:eval-unresolved}
gives the detailed aggregation, and Appendix~\ref{app:evaluation-coverage}
reports coverage, residual technical failure rates, and ranking sensitivity.
\label{sec:load-analysis}

\providecolor{CTBOrangeRow}{HTML}{FFF3E8}
\providecolor{CTBOrangeRowAlt}{HTML}{FFFAF5}
\section{Experiments}
\label{sec:experiments}

\label{sec:exp-settings}
Our experiments test whether TextCanvasBench separates current generators,
how performance changes as the required text and independently organized blocks
increase, and which failures text accuracy alone misses. We first compare model
scores and rankings across benchmarks, then stratify results by character and
text-block counts, inspect presentation and semantic errors, and isolate the
effect of benchmark inputs under a shared evaluator.
Nine generators (six open-weight, three proprietary) receive the same 1,600
prompts without reference images or evaluation specifications. We retain the
first valid output within a fixed retry budget and distinguish generation
failures from unresolved evaluation evidence. Dimension means use valid scores for that dimension; composite means use
images with valid scores in all four dimensions. Paired analyses retain
matched records across the conditions being compared.
Appendix~\ref{app:human-evaluation} describes the human rating protocol.
Appendix~\ref{app:evaluation-coverage} reports evaluation coverage, technical
failures, and sensitivity to unresolved checks.

\begin{table}[!t]
\centering
\caption{\textbf{Cross-benchmark comparison on a 0--1 scale.}
We compare published scores from CVTG~\citep{tai2026textcrafterv7},
LongText-Bench~\citep{geng2025xomni}, and OneIG-Bench~\citep{chang2025oneig}
with the English- and Chinese-prompt TextCanvasBench scores.
TextCanvasBench totals average the four dimensions within each image and then
average images with all four valid scores. Scores originally reported on a
0--100 scale are divided by 100.
Variance is
the population variance over the seven models shared by all columns; larger
values indicate greater cross-model dispersion. Spearman $\rho$ compares each
benchmark ranking with the human ranking over its available models: eight
for CVTG and OneIG, and nine for LongText and TextCanvasBench. English
human ratings are used for CVTG and OneIG; the language-specific columns use
the corresponding human ratings.
Dashes denote unavailable results.}
\label{tab:exp-cross-benchmark}
\label{tab:exp-summary-scores}
\label{tab:exp-main}
\begingroup
\ctbTableFont
\setlength{\tabcolsep}{3pt}
\renewcommand{\arraystretch}{1.20}
\begin{tabular*}{\linewidth}{@{\extracolsep{\fill}}l*{6}{c}@{}}
\toprule
\multicolumn{1}{c}{\multirow{2}{*}{\textbf{Generator}}}
& \textbf{CVTG}
& \multicolumn{2}{c}{\textbf{LongText}}
& \textbf{OneIG}
& \multicolumn{2}{c}{\textbf{Ours}} \\
\cmidrule(lr){2-2}
\cmidrule(lr){3-4}
\cmidrule(lr){5-5}
\cmidrule(l){6-7}
& Word & EN & ZH & Ovr.\ EN
& $S_{\mathrm{EN}}$ & $S_{\mathrm{ZH}}$ \\
\midrule
\ctbGroupRow{7}{Open-weight models}
FLUX.1 [dev]\ctbModelCite{bfl2024flux1dev}
& 0.4965 & 0.607 & 0.005 & 0.434  & 0.3839 & 0.2040 \\
SD3.5 Large\ctbModelCite{stability2024sd35large}
& 0.6548 & 0.4460 & 0.0079 & 0.462  & 0.2497 & 0.0565 \\
Qwen-Image\ctbModelCite{wu2025qwenimage}
& 0.8288 & 0.943 & 0.946 & 0.539  & 0.7013 & 0.7107 \\
Z-Image-Turbo\ctbModelCite{tongyi2025zimage}
& 0.8585 & 0.917 & 0.926 & 0.528  & 0.8002 & 0.8063 \\
GLM-Image\ctbModelCite{zai2026glmimage}
& 0.9116 & 0.952 & 0.979 & 0.528  & 0.7180 & 0.7289 \\
Boogu-Image-0.1-Turbo\ctbModelCite{chen2026booguimage}
& 0.819 & 0.944 & 0.977 & ---  & 0.7971 & 0.7987 \\
\addlinespace[2pt]
\rowcolor{CTBOrange}
\multicolumn{7}{@{}l}{\textcolor{white}{\bfseries Proprietary models}} \\
\rowcolor{CTBOrangeRow}
GPT-Image-2\ctbModelCite{openai2026gptimage2}
& 0.9003 & 0.964 & 0.983 & 0.516  & 0.9119 & 0.9135 \\
\rowcolor{CTBOrangeRowAlt}
Nano-Banana-Pro\ctbModelCite{google2025nanobananapro}
& 0.779 & 0.976 & 0.949 & 0.583  & 0.8850 & 0.8852 \\
\rowcolor{CTBOrangeRow}
Seedream-5.0-Pro\ctbModelCite{Seedream-5.0-Pro}
& --- & 0.941 & 0.946 & 0.553  & 0.8676 & 0.8714 \\
\midrule
\textbf{Variance} $\uparrow$
& 0.0195 & 0.0388 & 0.1846 & 0.0021 & 0.0547 & 0.1002 \\
\textbf{Human $\rho$} $\uparrow$
& 0.619 & 0.733 & 0.653 & 0.539 & 0.983 & 0.983 \\
\bottomrule
\end{tabular*}
\vspace{3pt}
\endgroup
\end{table}

\begin{table}[!t]
\centering

\caption{\textbf{TextCanvasBench diagnostic profiles.}
Dimension scores for the English- and Chinese-prompt tracks.
All scores are reported on a 0--100 scale. Each dimension averages its
available valid image scores; unresolved evaluations are not assigned zero.}

\label{tab:exp-diagnostic-profiles}

\begingroup
\ctbTableFont

\setlength{\tabcolsep}{3pt}
\renewcommand{\arraystretch}{1.20}

\begin{tabular*}{\linewidth}{
@{\extracolsep{\fill}}
l*{8}{c}
@{}
}

\toprule

\multicolumn{1}{c}{
    \multirow{2}{*}{\textbf{Generator}}
}
& \multicolumn{4}{c}{
    \textbf{English prompt}
}
& \multicolumn{4}{c}{
    \textbf{Chinese prompt}
} \\

\cmidrule(lr){2-5}
\cmidrule(l){6-9}

& D1
& D2
& D3
& D4
& D1
& D2
& D3
& D4 \\

\midrule

\ctbGroupRow{9}{Open-weight models}

FLUX.1 [dev]\ctbModelCite{bfl2024flux1dev}
& 70.97 & 12.84 & 51.46 & 17.10 & 59.74 & 8.06 & 12.25 & 4.36 \\

SD3.5 Large\ctbModelCite{stability2024sd35large}
& 50.06 & 5.90 & 36.42 & 7.76 & 15.75 & 1.46 & 6.90 & 1.01 \\

Qwen-Image\ctbModelCite{wu2025qwenimage}
& 84.21 & 39.46 & 90.23 & 61.34 & 85.68 & 41.24 & 90.45 & 61.75 \\

Z-Image-Turbo\ctbModelCite{tongyi2025zimage}
& 91.52 & 53.10 & 91.52 & 80.42 & 91.75 & 55.22 & 91.32 & 80.89 \\

GLM-Image\ctbModelCite{zai2026glmimage}
& 93.16 & 51.22 & 79.53 & 60.19 & 93.19 & 52.72 & 80.05 & 61.15 \\

Boogu-Image-0.1-Turbo\ctbModelCite{chen2026booguimage}
& 89.36 & 55.62 & 94.52 & 73.76 & 89.35 & 54.97 & 94.71 & 74.43 \\

\addlinespace[2pt]

\rowcolor{CTBOrange}
\multicolumn{9}{@{}l}{
    \textcolor{white}{\bfseries Proprietary models}
} \\

\rowcolor{CTBOrangeRow}
GPT-Image-2\ctbModelCite{openai2026gptimage2}
& 97.74 & 70.83 & 96.96 & 98.99 & 97.70 & 70.66 & 97.46 & 99.03 \\

\rowcolor{CTBOrangeRowAlt}
Nano-Banana-Pro\ctbModelCite{google2025nanobananapro}
& 95.47 & 66.14 & 94.93 & 97.02 & 95.47 & 66.44 & 94.95 & 97.26 \\

\rowcolor{CTBOrangeRow}
Seedream-5.0-Pro\ctbModelCite{Seedream-5.0-Pro}
& 94.73 & 64.07 & 95.56 & 91.34 & 95.02 & 64.27 & 95.80 & 92.04 \\

\bottomrule

\end{tabular*}
\endgroup
\end{table}

\subsection{Performance as Text Amount and Layout Complexity Increase}
\label{sec:exp-text-structure}

\paragraph{Cross-benchmark comparison.}
Table~\ref{tab:exp-cross-benchmark} reveals differences among generators that
receive similar scores on existing text-rendering benchmarks. GLM-Image and
GPT-Image-2 differ by only 0.004 on LongText Chinese, whereas their
TextCanvasBench Chinese scores differ by 0.1845. Relative model order also
depends on the benchmark: GLM-Image exceeds Z-Image-Turbo on both LongText
tracks, but Z-Image-Turbo obtains higher TextCanvasBench scores in both
languages. These comparisons motivate evaluating presentation and semantic
associations alongside textual correctness.

\paragraph{Agreement with human judgments.}
TextCanvasBench achieves a Spearman correlation of 0.983 with the human
composite ranking in both languages, as shown in
Table~\ref{tab:exp-cross-benchmark}. The corresponding correlations for the
comparison benchmarks range from 0.539 to 0.733. A prompt-cluster bootstrap
with 20,000 resamples gives a 95\% interval of $[0.950,0.983]$ for each
language, as reported in Table~\ref{tab:human-ranking-ci}. This agreement
concerns overall model ordering; the dimension-level profiles below explain
the distinct strengths underlying the composite scores.

\paragraph{Dimension-level profiles.}
Table~\ref{tab:exp-diagnostic-profiles} identifies the capabilities underlying
these differences. On the Chinese track, GPT-Image-2 exceeds GLM-Image by
4.51 points in D1, compared with 17.94 in D2, 17.40 in D3, and 37.87 in D4.
Their relatively close glyph-quality scores therefore coexist with larger
differences in content preservation, requested presentation, and semantic
alignment. The distinction is also apparent within individual models:
Qwen-Image scores 85.68 in D1 and 90.45 in D3, but 41.24 in D2. This gap reflects the distinction between producing well-formed lettering
in the requested style and reproducing the complete required content.
Z-Image-Turbo and Boogu-Image-0.1-Turbo show another trade-off: Boogu scores
higher in D3 on both tracks, whereas Z-Image-Turbo scores higher in D4.
The four-dimensional profile distinguishes these strengths rather than
reducing them to a single notion of text quality.

\begin{figure}[!t]
\centering
\includegraphics[width=\linewidth]{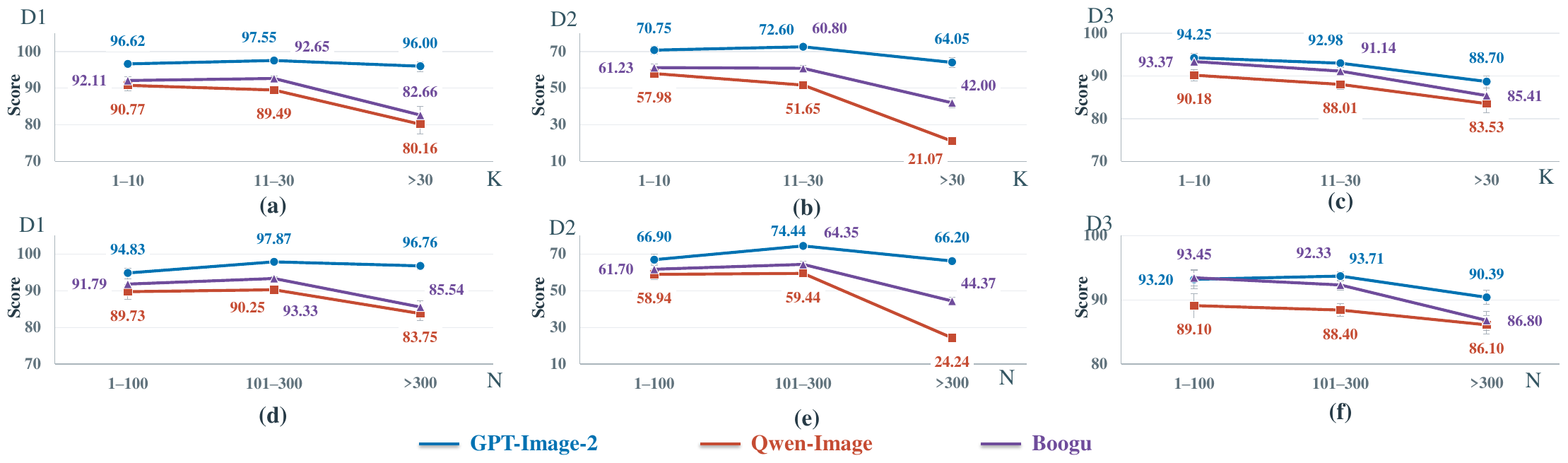}
\caption{\textbf{Performance as textual requirements increase.}
(a--c) D1--D3 by non-empty text-instance count $K$; (d--f) D1--D3 by required
character count $N$. Points are group means over common inputs across models
and dimensions; unavailable records are excluded jointly without zero
imputation.}
\label{fig:exp-text-structure}
\end{figure}

\paragraph{Text amount and layout complexity.}
Figure~\ref{fig:exp-text-structure} examines how D1--D3 vary with the number
of required text instances $K$ and required characters $N$. In panels (a--c),
Qwen-Image and Boogu exhibit pronounced D1 and D2 declines in the $K>30$
group, while GPT-Image-2 retains comparatively high glyph quality. D3 declines
across the three instance-count groups for all three generators, indicating
that satisfying presentation requirements becomes more difficult as more
text instances must be organized within an image.

Panels (d--f) show a related pattern for character count. The $N>300$ group
has lower D2 scores than the intermediate group for all three models, with
larger decreases for Qwen-Image and Boogu. D3 also declines between these
groups, while GPT-Image-2 remains comparatively stable in D1. The curves are
not uniformly monotonic: several scores increase from the shortest to the
intermediate group before falling in the longest group. Character count and
text-instance count thus provide complementary views of generation difficulty,
with the largest demands exposing errors that glyph quality alone does not
capture.

\subsection{What Text Accuracy Misses}
\label{sec:exp-errors}

\begin{figure}[!t]
\centering
\includegraphics[width=\linewidth]{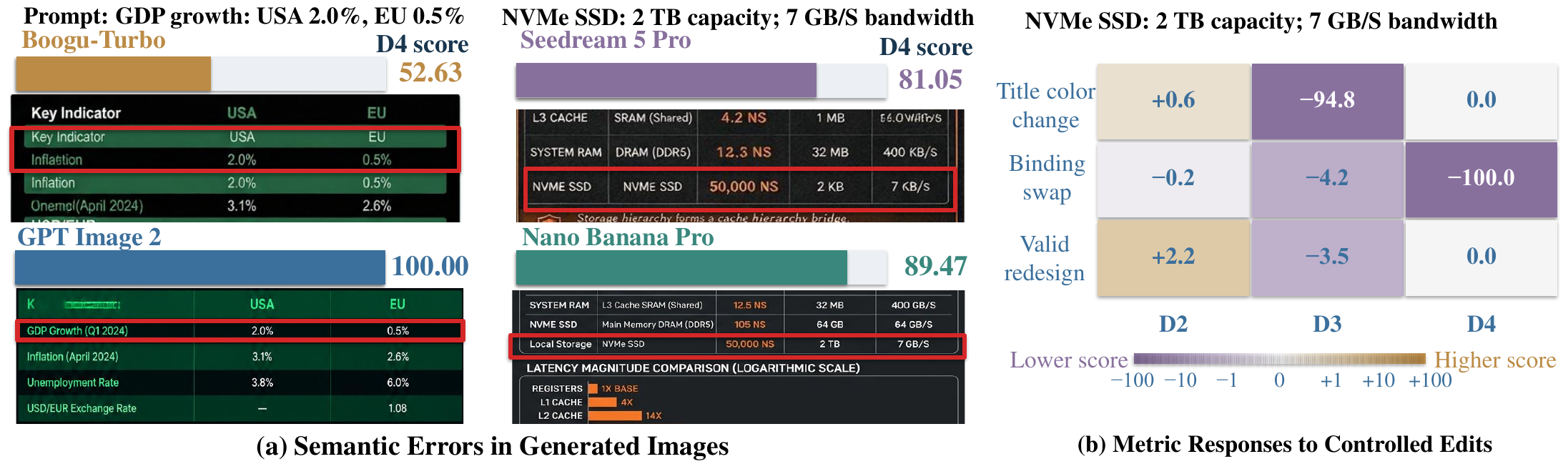}
\caption{\textbf{Semantic errors and responses to controlled edits.}
(a) Label--value and unit errors in generated images; red boxes identify the
discussed details. (b) Mean score changes relative to each unedited image for
title-color changes, text--object binding swaps, and valid redesigns.}
\label{fig:exp-diagnostics}
\end{figure}

Figure~\ref{fig:exp-diagnostics}a illustrates two limitations of text-only
evaluation. In the GDP example, Boogu renders the requested 2.0\% and 0.5\%
values but associates them with inflation rather than GDP growth; GPT-Image-2
preserves the intended association. Because the required strings remain
visible, text accuracy alone cannot identify this error. In the storage
example, Seedream-5.0-Pro changes 2~TB and 7~GB/s to 2~KB and 7~KB/s, whereas
Nano-Banana-Pro preserves both units. A text metric can detect these
substitutions, but D4 additionally captures their effect on the stated storage
capacity and bandwidth.
Figure~\ref{fig:exp-diagnostics}b tests evaluator responses using
24 reference images and three edits per image. Changing a specified title
color lowers the targeted D3 color score by 94.8 points without changing the
targeted D4 association score. Swapping existing text regions preserves
strings but changes associations: D2 changes by $-0.2$ points while the D4
association score falls by 100.0. Meaning-preserving spacing redesigns leave
the latter unchanged. The reported changes measure the targeted color and semantic-association
checks. Appendix~\ref{app:diagnostic-design}
provides the edit protocol and valid-pair counts.

\subsection{Ablation Study}
\begin{figure}[!htbp]
\centering
\includegraphics[width=\linewidth]{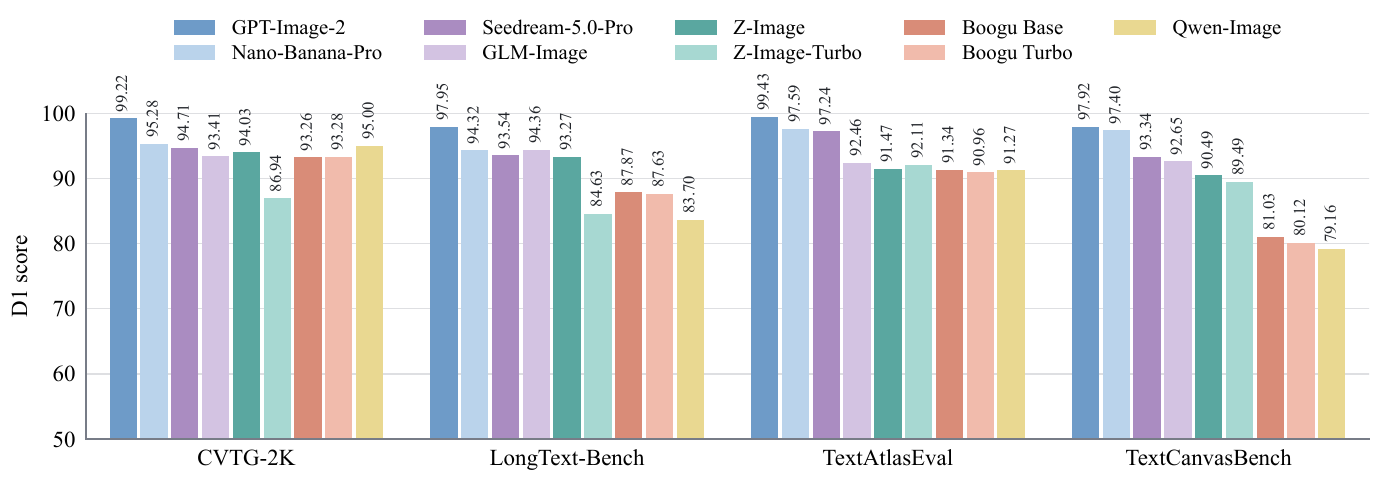}
\caption{\textbf{Cross-benchmark comparison with a shared evaluator.}
Nine generators are evaluated using the same TextPecker D1 protocol on
200 prompts per benchmark. TextCanvasBench bars average three independent
generations on identical prompts. Scores are on a 0--100 scale.}
\label{fig:exp-benchmark-ablation}
\end{figure}

To examine the effect of benchmark inputs while holding the evaluator fixed,
we apply the same full-image TextPecker D1 protocol to CVTG-2K,
LongText-Bench, TextAtlasEval, and TextCanvasBench. Each benchmark contributes
200 prompts shared across nine generators. The comparison excludes FLUX.1
and SD3.5 and includes both base and Turbo variants of Boogu and Z-Image,
so separation is not driven solely by the weakest baselines. The
TextCanvasBench subset emphasizes longer required text; its bars average
three independent generations for each prompt.

Figure~\ref{fig:exp-benchmark-ablation} shows a cross-model D1 range of
18.76 points on TextCanvasBench, compared with 12.28 on CVTG-2K, 14.25 on
LongText-Bench, and 8.47 on TextAtlasEval. The corresponding population
variances of the nine model means are 47.10, 9.06, 21.88, and 9.83.
TextCanvasBench therefore retains greater model separation under a common
scoring protocol in this comparison. Within model families, Z-Image exceeds
Z-Image-Turbo by 0.99 points and Boogu Base exceeds Boogu Turbo by 0.91
points on TextCanvasBench. The latter difference is $-0.02$, $+0.24$, and
$+0.38$ points on the three comparison benchmarks, respectively.
Appendix~\ref{app:ranking-robustness} examines score and rank stability
across the three independent generations.

\begingroup
\subsection{Content Preservation and Model-Design Implications}
\label{sec:exp-design-insights}

\paragraph{Field identity and semantic associations beyond glyph quality.}
Readable text need not preserve the required information.
In Figure~\ref{fig:additional-content-errors}b, Seedream's menu retains the same
required item, price, unit, and requested container count, but expanding
surrounding text from 256 to 1024 target characters changes RMB~12 per skewer
to RMB~12 per portion. Boogu must assign Brown Rice 37\%, yet assigns it both
43\% and 37\%. These are field substitutions and conflicting associations,
not merely unreadable glyphs. Across its 192 audited core instances, Boogu
has 81 missing/substituted instances and 12 definite binding errors/conflicts;
15 of its 21 outputs with available $D1\ge90$ still contain a readable
substitution or definite binding error, as shown in Figure~\ref{fig:additional-content-errors}a.
The controlled edits in Figure~\ref{fig:exp-diagnostics}b separately probe
these distinctions.

\begin{figure}[!t]
\centering
\includegraphics[width=\linewidth]{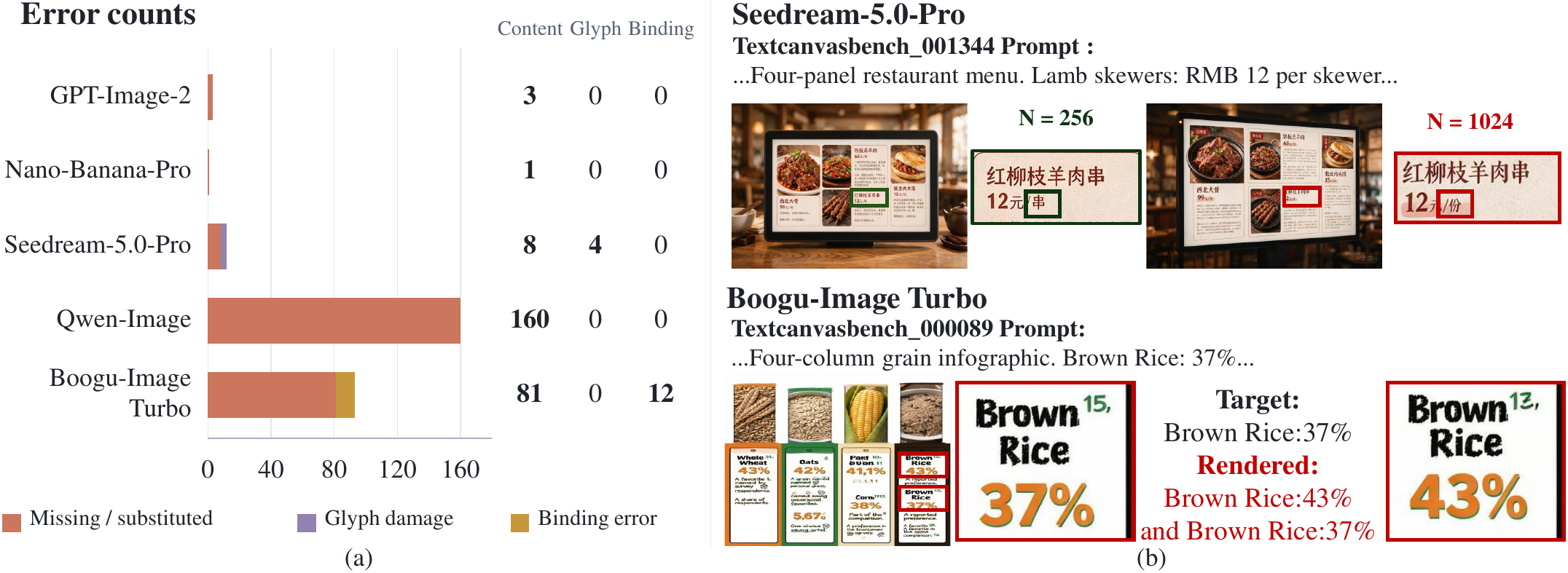}
\caption{Core-information errors and examples across 240 existing outputs. Each of five models contributes 48 images and 192 predefined core-fact instances. Panel a groups confirmed errors; unresolved evidence is not classified as an error. Panel b shows Seedream's readable unit substitution as text increases and Boogu's conflicting Brown Rice associations.}
\label{fig:additional-content-errors}
\end{figure}

The public model designs summarized in Table~\ref{tab:additional-model-design}
include glyph encoding,
joint text--visual processing, paragraph/template curricula, and character
coverage. These capabilities need to be coupled with full-text identity and
relations: the observed errors motivate representations retaining text-instance
identities and entity--value--unit bindings, supervision checking substitutions
and contradictions, and rendering that preserves field correspondence rather
than only the presence of strings.

\paragraph{Unchanged core information under increased surrounding text.}
We analyze 240 existing outputs: 12 scenes (six per language), five models,
and target character counts $N\in\{256,1024\}$ crossed with requested
container counts $K\in\{4,16\}$. Four predefined core facts remain fixed per
scene. At matched scene, model, $K$, and actual resolution, increasing $N$
reduces core-information fidelity by 12.6 points for Seedream and 22.9 for
Boogu, with exploratory 95\% scene-bootstrap intervals of $[-24.0,-3.4]$ and
$[-40.8,-4.4]$, respectively, as shown in
Figure~\ref{fig:additional-paired-effects}a.
GPT-Image-2 and Nano-Banana-Pro change by $-3.1$ and $+1.0$ points and retain
189 and 191 of 192 audited instances. Tracking the same facts distinguishes
damage to existing fields from errors confined to newly added text.

This suggests paired training data with fixed facts and expanded surrounding
text, supervising both added-text accuracy and unchanged-field retention.
Planning and layout assignment can preserve field identities; rendering and
subsequent corrections can recheck values, units, and associations to avoid
omissions or conflicts during content expansion. These design directions
follow from the observed content-retention errors.
Appendix~\ref{app:diagnostic-design} gives the complete protocol,
results, and public-model-design comparisons.
\endgroup

\section{Related Work}
\label{sec:related-work}
\paragraph{Text-rich generation benchmarks.}
TextAtlasEval studies dense text generation~\citep{wang2025textatlas5m},
CVTG-2K evaluates multiple text instances in complex scenes~\citep{du2025textcrafter},
and LongText-Bench tests extended bilingual text rendering~\citep{geng2025xomni}.
TextGround4M emphasizes prompt--layout alignment~\citep{mao2026textground4m}.
OneIG-Bench covers multiple generation dimensions~\citep{chang2025oneig},
while BizGenEval examines text, layout, attribute binding, and reasoning in
commercial visuals~\citep{li2026bizgeneval}. TextCanvasBench combines
reference-grounded tasks across eight domains with stable text-block identities
and separate presentation and semantic specifications. Character and block
counts enable complementary analyses of textual and structural load.
Appendix~\ref{app:cb-comparison} compares benchmark coverage and annotation scope.
\textbf{Diagnostic evaluation.}
TextPecker quantifies structural glyph anomalies~\citep{zhu2026TextPecker}.
For semantic evaluation, LongT2IBench uses entity--attribute--relation
annotations~\citep{yang2026longt2ibench}, while IGenBench verifies
infographic requirements through atomic questions~\citep{tang2026igenbench}.
Building on these directions, our protocol separates glyph integrity,
content correctness, layout/style fidelity, and semantic alignment.
Frozen specifications define shared targets across generators; explicit
evidence states distinguish confirmed violations from unresolved observations.
Controlled edits further probe whether presentation and association changes
affect the intended diagnostic dimensions.

\section{Conclusion}
\label{sec:conclusion}
We introduced \textsc{TextCanvasBench}, a benchmark of 1,600 specification-guided samples across eight domains and 38 subcategories of practical text-rich visuals. Its four dimensions assess glyph quality, complete-text accuracy, prompt-specified layout and style, and semantic alignment from the whole image to text blocks and regions. Across nine open- and closed-source generators, the benchmark distinguishes models with similar scores on existing benchmarks, and its rankings agree more closely with human judgments than those of most benchmarks in our comparison.

\bibliography{iclr2026_conference}
\bibliographystyle{iclr2027_conference}

\clearpage
\appendix

\section{Dataset Statistics and Benchmark Comparison}
\label{app:cb-data-statistics}

We characterize each sample by its required-text length, text instances, and
functional regions, as summarized in Table~\ref{tab:cb-text-detail}. A text instance is a
non-empty entry in the required-text inventory; repeated strings at distinct
occurrences are counted separately. Related instances are organized into
functional regions, such as a title area, paragraph, or table. The 1,600 samples
contain 43,390 text instances in 12,399 regions, averaging 27.12 instances and
7.75 regions per image.

Required-text length $N$ counts non-whitespace characters, including punctuation,
across the text instances. Scene descriptions, region identifiers, and table
formatting delimiters are excluded. The instance count $K$ and all length-based
cross-tabulations use this same inventory.

\begin{table}[!htbp]
\centering\small
\caption{\textbf{Macro- and micro-level dataset statistics.} All counts use the
1,600-input inventory. The two right-hand groups are separate cross-tabulations
of the same samples; $K$ bins are not layout levels. P90 denotes the
90th percentile, computed by linear interpolation.}
\label{tab:cb-text-detail}
\label{tab:cb-length-layout-joint}
\label{tab:cb-length-block-joint}
\begingroup
\setlength{\tabcolsep}{5pt}
\renewcommand{\arraystretch}{1.05}
\begin{tabular*}{\linewidth}{@{\extracolsep{\fill}}lrrrr@{}}
\hline
Per-image quantity & Mean & Median & P90 & Range \\
\hline
Characters $N$ & 633.73 & 240 & 1,630.20 & 4--8,423 \\
Text instances $K$ & 27.12 & 17 & 65.10 & 1--248 \\
Functional regions $R$ & 7.75 & 7 & 13 & 1--59 \\
\hline
\end{tabular*}
\par\vspace{5pt}
\begin{tabular*}{\linewidth}{@{\extracolsep{\fill}}lrrrrrrr@{}}
\hline
 & & \multicolumn{3}{c}{Layout level} & \multicolumn{3}{c}{Text instances $K$} \\
\cline{3-5}\cline{6-8}
Characters $N$ & Samples & L1 & L2 & L3 & 1--10 & 11--30 & $>30$ \\
\hline
1--100 & 261 & 76 & 108 & 77 & 210 & 51 & 0 \\
101--300 & 657 & 107 & 293 & 257 & 185 & 423 & 49 \\
$>300$ & 682 & 46 & 203 & 433 & 50 & 224 & 408 \\
\hline
Total & 1,600 & 229 & 604 & 767 & 445 & 698 & 457 \\
\hline
\end{tabular*}
\endgroup
\end{table}

\subsection{Domain and Subcategory Coverage}
\label{app:cb-category-details}

The eight domains expose different combinations of textual content and visual
organization, as shown in Table~\ref{tab:cb-domain-profiles}. Academic documents and slides
contain the most required text on average, whereas infographics have the
highest median instance count and L3 share. Digital interfaces contain a median
of 27 text instances and a mean of 328.8 characters, reflecting their dense
use of short labels and values. Together, these profiles show that the same
amount of required text can be organized differently across practical visual
media.

\begin{table}[!htbp]
\centering\small
\caption{\textbf{Domain-level text and layout profiles.} Long-text and L3 percentages use
the corresponding domain count as denominator.}
\label{tab:cb-domain-profiles}
\begingroup
\setlength{\tabcolsep}{4pt}
\renewcommand{\arraystretch}{1.05}
\begin{tabular*}{\linewidth}{@{\extracolsep{\fill}}lrrrrr@{}}
\hline
Domain & Samples & Mean chars. & Median instances & $N>300$ (\%) & L3 (\%) \\
\hline
Infographics & 310 & 1323.5 & 58 & 95.48 & 89.35 \\
Posters \& Ads & 265 & 260.5 & 15 & 24.15 & 45.28 \\
Art \& Illustration & 215 & 296.6 & 10 & 21.86 & 46.51 \\
Document Design & 225 & 727.9 & 13 & 37.33 & 19.56 \\
Real-world Photos & 160 & 206.2 & 11.5 & 17.50 & 32.50 \\
Spatial Signage & 160 & 217.5 & 9.5 & 19.38 & 35.62 \\
Digital Interfaces & 170 & 328.8 & 27 & 42.94 & 51.76 \\
Academic \& Slides & 95 & 1930.7 & 21 & 62.11 & 30.53 \\
\hline
\end{tabular*}
\endgroup
\end{table}

Table~\ref{tab:cb-subcategory-counts} expands the eight domains into 38
concrete content topics and use cases. Domain-specific subcategory counts range
from two for Academic \& Slides to nine for Infographics, while individual
subcategories contain 24--84 samples. This hierarchy preserves broad visual
media coverage while preventing the benchmark from collapsing into a single
content type within each domain.

\begin{table*}[!t]
\centering
\caption{\textbf{Taxonomy and sample counts across 38 subcategories.}
Parentheses report subcategory counts; the final column reports domain totals.
Each of the 1,600 samples belongs to exactly one subcategory.}
\label{tab:cb-subcategory-counts}
\begingroup
\footnotesize
\setlength{\tabcolsep}{5pt}
\renewcommand{\arraystretch}{1.10}
\begin{tabularx}{\textwidth}{@{}>{\raggedright\arraybackslash}p{.17\textwidth}>{\raggedright\arraybackslash}Xr@{}}
\hline
\textbf{Domain} & \textbf{Subcategories (samples)} & \textbf{Total} \\
\hline
Infographics & Society \& humanities (29); Geography (35); Data charts (29); Life \& health (29); Everyday guides (29); Science \& engineering (52); Economy \& business (28); Natural sciences (47); Arts \& culture (32). & 310 \\
Posters \& Ads & Public-service posters (60); Commercial ads (41); Film \& TV posters (49); Recruitment posters (43); Event posters (72). & 265 \\
Art \& Illustration & Traditional painting (41); Illustrations (84); Comics \& anime (46); Modern collage (44). & 215 \\
Document Design & Newspaper pages (26); Resumes (42); Thank-you letters (40); Holiday greetings (46); Invitations (71). & 225 \\
Real-world Photos & Meetings (24); Event scenes (37); Objects \& products (50); Everyday settings (25); Natural settings (24). & 160 \\
Spatial Signage & Traffic signage (44); Public facilities (33); Commercial spaces (50); Industrial signage (33). & 160 \\
Digital Interfaces & Apps (44); Game interfaces (38); Web pages (25); Software interfaces (63). & 170 \\
Academic \& Slides & Presentation slides (50); Paper pages (45). & 95 \\
\hline
\textbf{Total} & \textbf{38 subcategories across eight domains} & \textbf{1,600} \\
\hline
\end{tabularx}
\endgroup
\end{table*}

\begin{samepage}
\subsection{Comparison with Related Benchmarks}
\label{app:cb-comparison}

Table~\ref{tab:cb-category-comparison} positions TextCanvasBench among
benchmarks for visual text generation and related compositional tasks.
Our design combines practical task coverage with text-block identities,
functional roles, explicit presentation requirements, and atomic semantic
questions. These annotations support evaluation from the whole image down to
individual blocks, separating glyph structure quality, textual content
correctness, layout and style, and text--image semantic alignment that
aggregate benchmark scores may obscure.
\par
\end{samepage}

\begin{table}[!htbp]
\centering
\caption{\textbf{Benchmark annotations and evaluation design.}
$\checkmark$: supported; \textemdash{}: unverified; N/A: outside the stated task.
Unknown entries do not indicate absence. Positions refer to annotations,
not evaluator-predicted boxes; reference images need not be generation inputs.}
\label{tab:cb-category-comparison}
\label{tab:cb-layout-comparison}
\label{tab:cb-spatial-comparison}
\begingroup
\fontsize{7.5}{8.5}\selectfont
\setlength{\tabcolsep}{2pt}
\renewcommand{\arraystretch}{1.18}
\renewcommand{\tabularxcolumn}[1]{m{#1}}
\begin{tabularx}{\linewidth}{@{}>{\raggedright\arraybackslash}m{.25\linewidth}*{7}{>{\centering\arraybackslash}X}>{\centering\arraybackslash}m{.135\linewidth}@{}}
\hline
 & \multicolumn{5}{c}{\textbf{Annotations}} & \multicolumn{3}{c}{\textbf{Evaluation design}} \\
\cline{2-6}\cline{7-9}
\multicolumn{1}{c}{\textbf{Benchmark}} & \shortstack{\textbf{Text}\\\textbf{instances}} & \textbf{Position} & \textbf{Style} & \textbf{Roles} & \shortstack{\textbf{Ref.}\\\textbf{images}} & \shortstack{\textbf{Joint}\\\textbf{strata}} & \shortstack{\textbf{Explicit}\\\textbf{checks}} & \shortstack[c]{\textbf{Semantic}\\\textbf{method}} \\
\hline
TextAtlasEval \citep{wang2025textatlas5m} & \textemdash{} & \textemdash{} & \textemdash{} & \textemdash{} & $\checkmark$ & \textemdash{} & \textemdash{} & CLIP \\
CVTG-2K \citep{du2025textcrafter} & $\checkmark$ & Text & $\checkmark$ & \textemdash{} & \textemdash{} & \textemdash{} & \textemdash{} & CLIP / VQA \\
LongText-Bench \citep{xomni2025repository} & \textemdash{} & \textemdash{} & \textemdash{} & \textemdash{} & \textemdash{} & \textemdash{} & \textemdash{} & \textemdash{} \\
OneIG-Bench \citep{chang2025oneig} & \textemdash{} & \textemdash{} & \textemdash{} & \textemdash{} & \textemdash{} & \textemdash{} & \textemdash{} & VLM \\
TextGround-Bench \citep{mao2026textground4m} & $\checkmark$ & Bbox & \textemdash{} & \textemdash{} & $\checkmark$ & $\checkmark$ & \textemdash{} & CLIP \\
BizGenEval \citep{li2026bizgeneval} & \textemdash{} & Text & $\checkmark$ & \textemdash{} & $\checkmark$ & \textemdash{} & $\checkmark$ & Checklist VQA \\
IGenBench \citep{tang2026igenbench} & \textemdash{} & \textemdash{} & \textemdash{} & \textemdash{} & $\checkmark$ & \textemdash{} & $\checkmark$ & Atomic VQA \\
\hline
LongT2IBench \citep{yang2026longt2ibench} & N/A & N/A & N/A & N/A & \textemdash{} & N/A & $\checkmark$ & Graph-based VLM \\
SpatialGenEval \citep{wang2026spatialgeneval} & N/A & N/A & N/A & N/A & \textemdash{} & N/A & \textemdash{} & Spatial MCQ \\
OverLayBench \citep{li2025overlaybench} & N/A & N/A & N/A & N/A & $\checkmark$ & N/A & N/A & \textemdash{} \\
\hline
\textbf{TextCanvasBench (ours)} & $\checkmark$ & Text & $\checkmark$ & $\checkmark$ & $\checkmark$ & $\checkmark$ & $\checkmark$ & Atomic VQA \\
\hline
\end{tabularx}
\endgroup
\end{table}

The comparison separates text-instance annotations from the object layouts
used in adjacent tasks. Joint strata retain each benchmark's native text-unit
definition, while explicit checks include prompt-derived questions and graph
criteria; neither implies an identical evaluation protocol. OneIG's semantic
entry refers to its broader evaluation framework.

\section{D3 and D4 Specification Preparation}
\label{app:cb-annotation}

This appendix details the construction process in
Section~\ref{sec:d3_d4_spec}. Final generation inputs define the requirements;
reference annotations assist text identification and grouping. Specification
construction precedes generated-image evaluation and does not use generator
outputs to select requirements.

\subsection{Generation Prompt Construction}
\label{app:cb-rubrics}

Each generation input combines a Chinese or English scene description with
the same original-language required-text inventory and rendering instructions.
The descriptions express matching content and presentation requirements.
The inventory groups text instances into functional regions while retaining
the wording, capitalization, numbers, punctuation, and repeated occurrences.
Each table preserves row membership and column correspondence. Structural
identifiers and table delimiters guide interpretation and are not additional
strings to render. The generator receives the assembled text without the
reference image.

Reference annotations support the association between text instances and
regions. Properties enter a specification only when supported by the final
input. For example, ``in the top left,'' ``below,'' and ``white text'' support
placement, order, and color checks respectively; unmentioned reference-image
appearance does not supply additional targets.

\begin{figure}[!htbp]
\centering
\includegraphics[width=\linewidth]{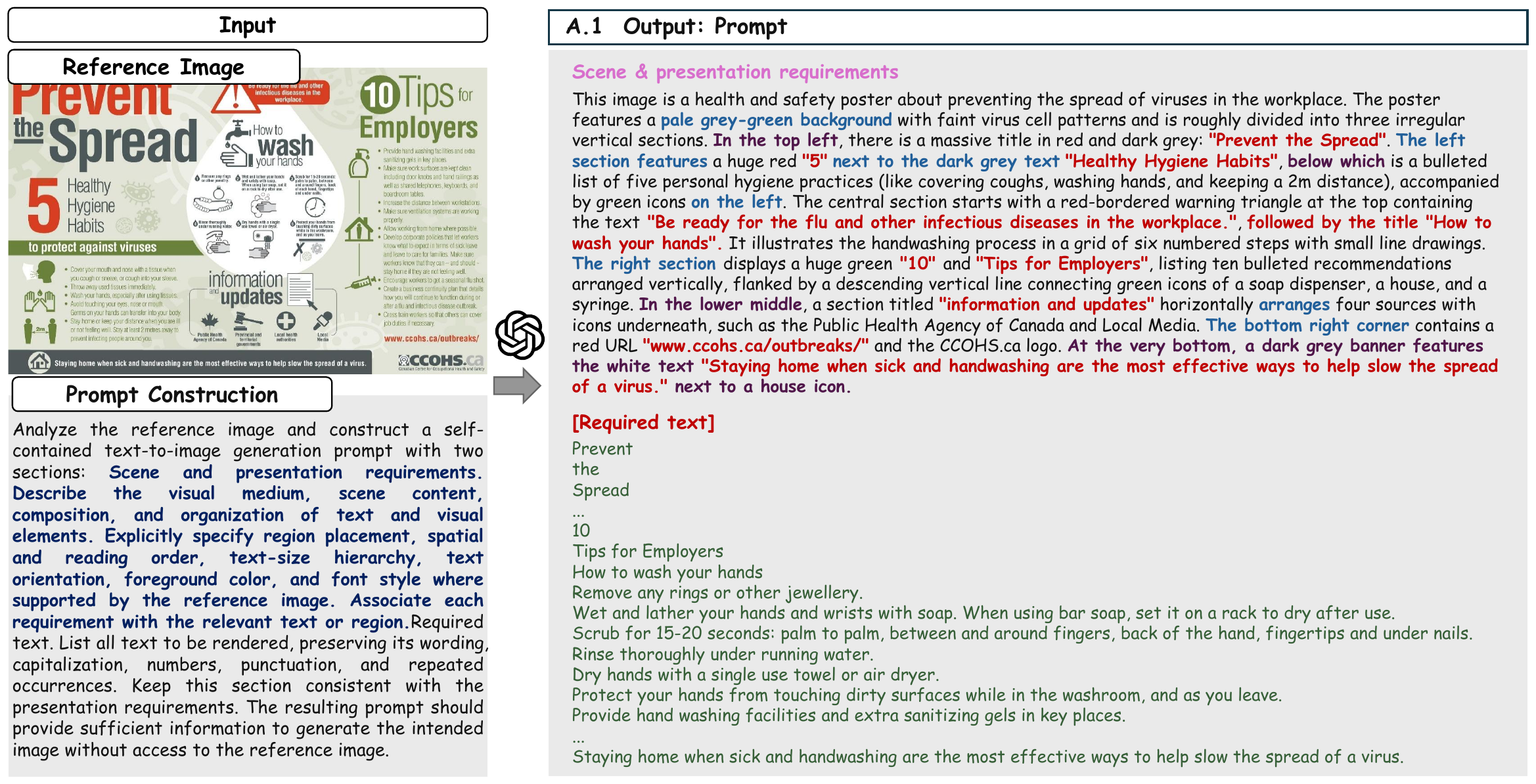}
\caption{\textbf{Generation prompt construction.} Reference-image content is
organized into a scene description and required-text inventory. The displayed
text is excerpted; the complete input retains the required strings.}
\label{fig:app-generation-prompt}
\end{figure}

\subsection{Structured Presentation Specifications (D3)}
\label{app:cb-prompt-spec}

D3 construction first extracts explicit text-presentation requirements from
the final prompt. Each requirement is assigned to placement, spatial/reading
order, size hierarchy, orientation, foreground color, or font style. Source
review checks the intended meaning and separates requirements on different
attributes. General descriptions of the scene are not converted into
unstated text-level constraints.

Requirements are then associated with identified text instances, groups, or
functional regions. Distinct occurrences of the same string retain separate
identities. Quantifiers such as each, all, some, and global remain part of the
requirement: a condition on every heading differs from one requiring at least
one qualifying heading. Conditional requirements retain their selection
conditions and exceptions. Grouping a heading with its body text does not
assign the heading's color or size to the entire group.

Each released entry contains \texttt{check\_id}, \texttt{requirement},
\texttt{targets}, \texttt{quantifier}, and \texttt{criterion}. Criteria describe
expected attributes, relations, or numerical ranges. Source quotations and
revision reasons are retained in construction records, separately from the
compact release. Structural checks validate identifiers and field consistency;
semantic review checks source support, target association, duplication, and
omitted requirements. The final inventory contains 33,262 checks, shared by
the Chinese and English descriptions of the 1,600 samples.

\begin{figure}[!tp]
\centering
\includegraphics[width=\linewidth]{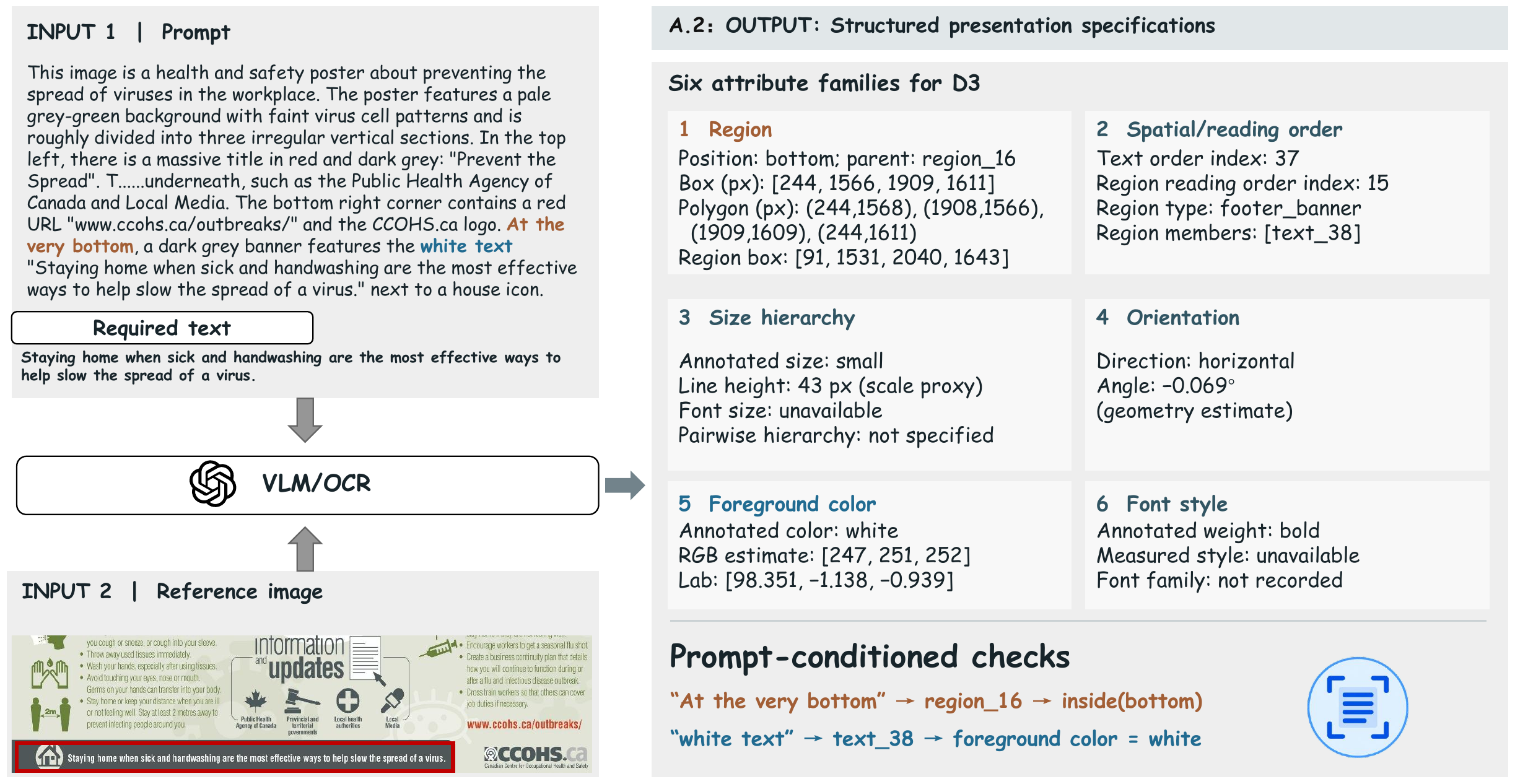}
\caption{\textbf{Example of D3 specification construction.}
The six attribute panels present reference annotations. The bottom panel
uses placement and color requirements to illustrate the mapping from prompt
requirements to text targets and evaluation criteria.}
\label{fig:layout-style-evidence-construction}
\end{figure}

Figure~\ref{fig:layout-style-evidence-construction} illustrates D3
specification construction. Text identities, region membership, and reference
annotations across the six attribute families support the association of
prompt requirements with their intended targets and the formulation of
structured checks. The footer banner's placement and text color provide
examples of this mapping.

\subsection{Fine-Grained Semantic Specifications (D4)}
\label{app:cb-d4-spec}

\begin{figure}[!tp]
\centering
\includegraphics[width=\linewidth]{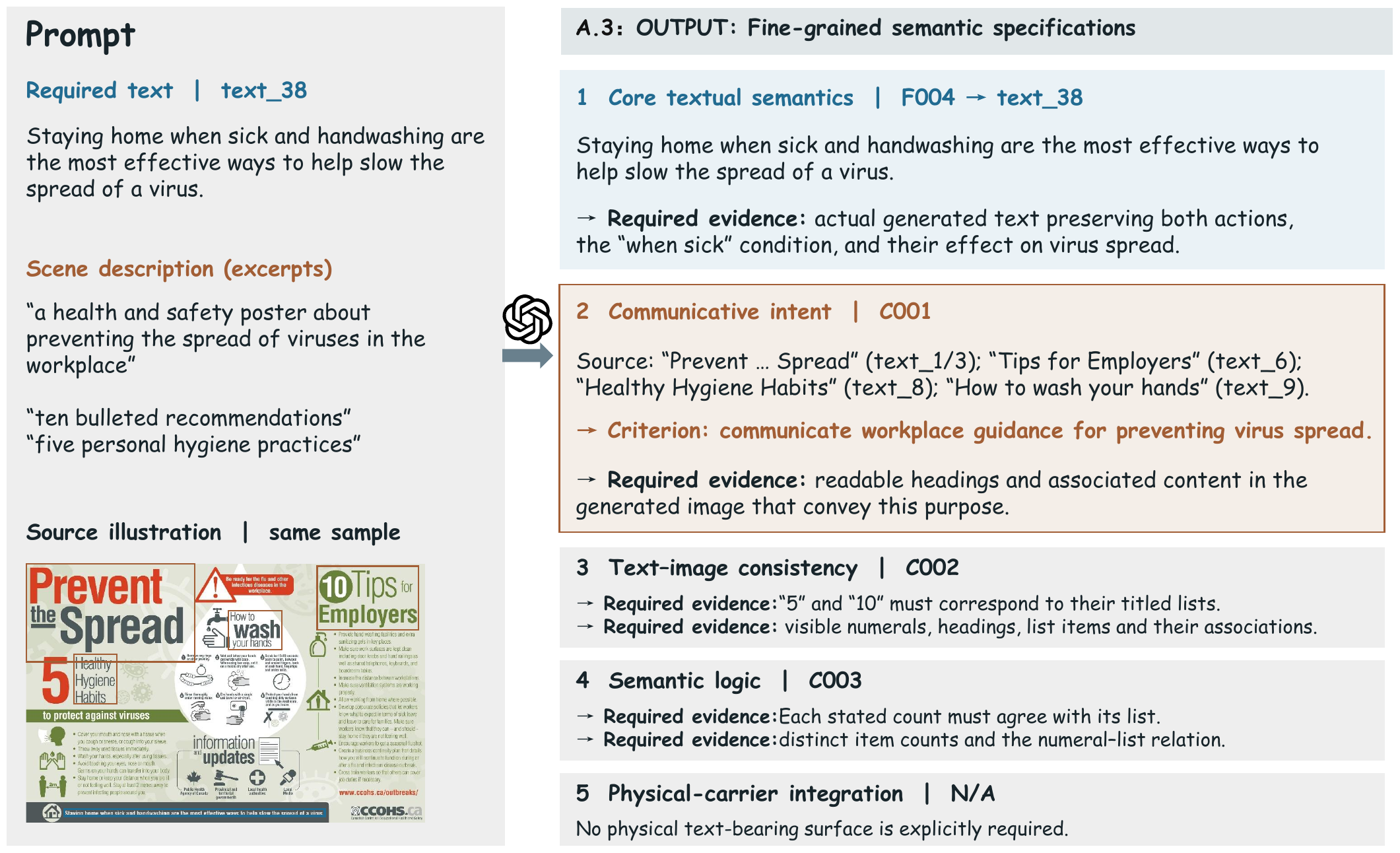}
\caption{\textbf{Example of D4 specification construction.} A workplace
health poster illustrates how required text and prompt context inform checks
across the five semantic components and their evidence requirements.
Physical-carrier integration is not applicable to this example.}
\label{fig:app-d4-construction}
\par\vspace{0.8em}
\includegraphics[width=.95\linewidth]{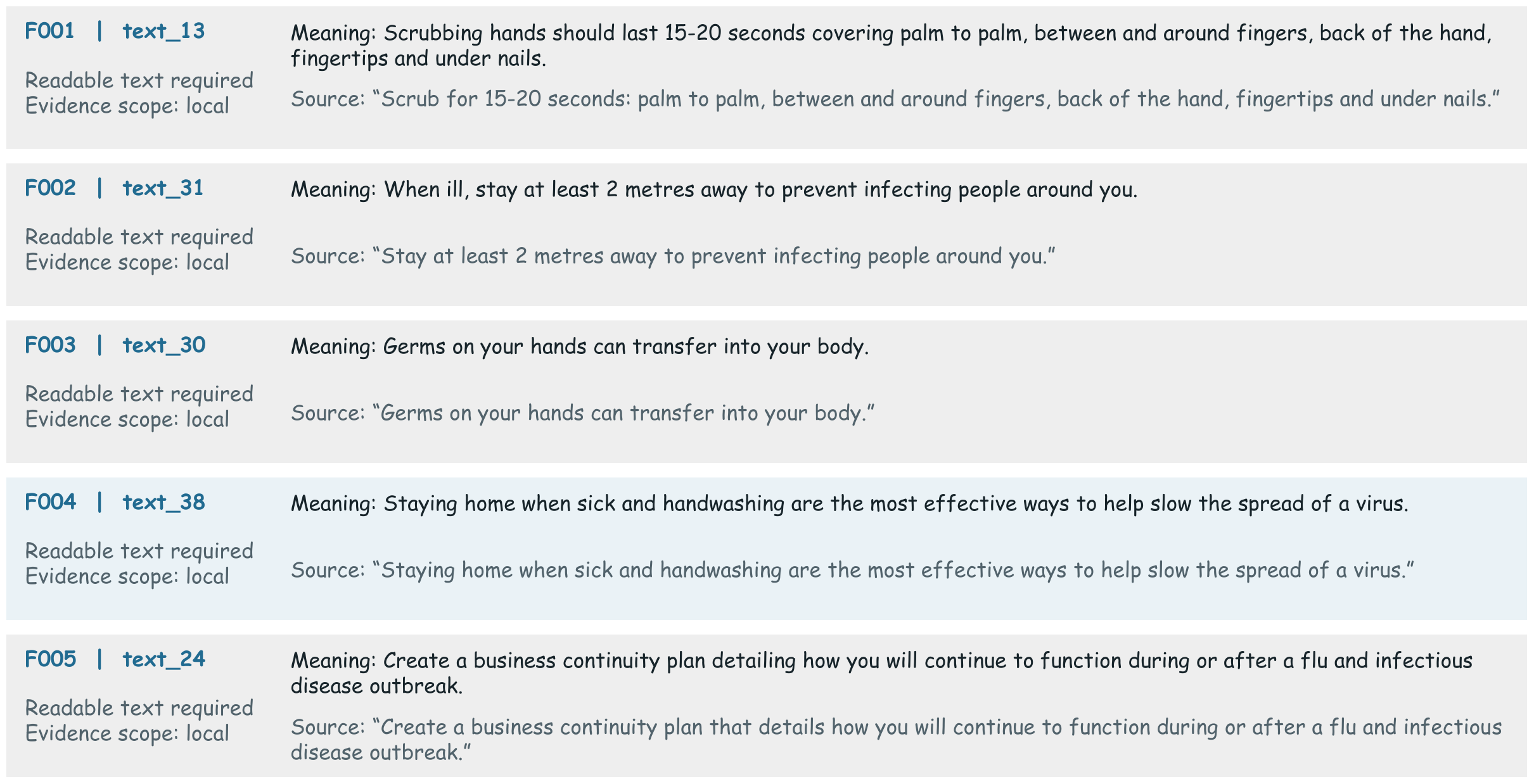}
\caption{\textbf{Illustrative core-semantic checks.} Five examples from the
poster in Figure~\ref{fig:app-d4-construction} link semantic requirements to
text targets and source passages, retaining the relevant actions, values,
units, and conditions.}
\label{fig:semantic-spec-evidence}
\end{figure}

D4 specifications are constructed from the final prompt, required text, and
region structure. A VLM drafts semantic checks, which are reviewed for source
support, granularity, and target association. Figure~\ref{fig:app-d4-construction}
illustrates the five components using a workplace health poster. Core textual
semantics preserves the meaning of the required statements; communicative
intent captures the poster's preventive guidance; text--image consistency
associates the numerals with their titled lists; and semantic logic checks
whether the stated counts agree with the list contents. Physical-carrier
integration applies when the prompt specifies a text-bearing object and
concerns attachment to its surface, perspective or curvature, and material
integration. It is not applicable in the illustrated example. The number of
checks depends on the supported requirements, with no fixed fact-count limit.

Figure~\ref{fig:semantic-spec-evidence} presents five illustrative core-semantic
checks from the same poster, linking required meanings to text targets and
source passages. Each check expresses an independently assessable requirement:
independent assertions are separated, while subjects, values, units, conditions,
negations, and referents remain bound where necessary to preserve meaning.
For example, the hand-scrubbing instruction retains its duration and specified
hand areas, while the footer statement retains both preventive actions and
the condition ``when sick.'' The same principle applies to tables, where a
fact binds a row subject to its column attribute, value, and unit. Equivalent
repetitions are merged, while distinct referents and occurrence-specific
relations remain separate. Meaning-preserving paraphrases are allowed; literal
text accuracy is evaluated by D2, and presentation requirements by D3.

Each check records its source, target identifiers, semantic requirement,
pass/fail criteria, and the evidence to inspect in a generated image.
Propositions are assigned a primary component to avoid duplicate evaluation
of the same fact. These specifications define the assessment requirements;
actual observations are collected during evaluation. Review records retain
source ambiguities, exclusions, and applicability reasons. The released
specifications contain 39,387 checks, including 28,914 core-semantic checks;
1,476 of the 1,600 samples contain more than five core-semantic checks.

\subsection{Semantic Fact Coverage}
\label{app:d4-coverage-extension}

Construction seeks broad coverage of substantive input meaning while retaining
source support for each check. We quantify the resulting scope using a
separately extracted and semantically reviewed inventory of necessary facts
from the same final inputs. This automated audit preserves subjects, values,
units, conditions, negations, and referents, and deduplicates equivalent facts
within an input. Incomplete source fragments are recorded separately rather
than completed by inference. The Chinese and English descriptions share the
required lettering, so each input is counted once.

Two separate semantic mappings associate the checklist with this inventory.
A fact is counted as covered only when both mappings identify checks that
cover its complete meaning. Coverage is assessed fact by fact within each region. Necessary facts
excluded during checklist construction remain in the audit denominator. The pooled coverage is
\begin{equation}
 C_{\mathrm{fact}}=
 \frac{\sum_i |\mathcal F_i^{\mathrm{covered}}|}
 {\sum_i |\mathcal F_i|},
\end{equation}
where $\mathcal F_i$ is the audited necessary-fact inventory for input $i$.
Across 1,600 inputs, the checklist covers 36,822 of 45,086 necessary facts
or 81.67\%, as shown in Table~\ref{tab:d4-fact-coverage}. This measures the semantic scope
of the checklist, separately from whether its checks receive determinate
judgments during generated-image evaluation.

\begin{table}[!htbp]
\centering\small
\caption{\textbf{Necessary-fact coverage of the released D4 checklist.}
Facts are deduplicated within each input; the total percentage uses pooled
fact counts.}
\label{tab:d4-fact-coverage}
\setlength{\tabcolsep}{4pt}
\begin{tabularx}{\linewidth}{@{}Xrrrr@{}}
\hline
Domain & Inputs & Necessary facts & Covered & Coverage (\%) \\
\hline
Infographics & 310 & 16,098 & 13,150 & 81.69 \\
Academic documents / slides & 95 & 3,938 & 2,643 & 67.12 \\
Real-world photos & 160 & 2,362 & 2,179 & 92.25 \\
Digital interfaces & 170 & 4,505 & 4,008 & 88.97 \\
Document design & 225 & 6,956 & 4,781 & 68.73 \\
Posters / ads & 265 & 4,902 & 4,521 & 92.23 \\
Spatial signage & 160 & 2,742 & 2,401 & 87.56 \\
Art / illustration & 215 & 3,583 & 3,139 & 87.61 \\
\textbf{Total} & 1,600 & 45,086 & 36,822 & 81.67 \\
\hline
\end{tabularx}
\end{table}

\subsection{Controlled Semantic-Error Detection}

We compare a restricted five-entry checklist with the full D4 checklist on
the same 24 pairs of typeset images. Each pair contains a correct image and
a variant with one inserted core-semantic error. Shared facts are checked
under both settings; added facts are checked only by the full D4 checklist.
A pair is eligible if both settings pass the correct image. An error counts
as detected when the judgment evidence identifies the inserted error.
One shared omission pair was excluded because it did not meet the eligibility
criterion.

\begin{table}[!htbp]
\centering\small
\caption{\textbf{Controlled semantic-error detection.} Tested and eligible
counts are reported separately. Detection counts use eligible pairs as the
denominator.}
\label{tab:d4-error-detection}
\setlength{\tabcolsep}{4pt}
\begin{tabular}{llrrrr}
\toprule
Scope & Error type & Tested & Eligible & Restricted (five entries) & Full D4 checklist \\
\midrule
Shared & Omission & 3 & 2 & 2/2 & 2/2 \\
Shared & Value / unit & 3 & 3 & 3/3 & 3/3 \\
Shared & Object association & 3 & 3 & 3/3 & 3/3 \\
Shared & Condition / negation & 3 & 3 & 3/3 & 3/3 \\
\textbf{Shared total} & --- & 12 & 11 & 11/11 & 11/11 \\
Added & Omission & 3 & 3 & 0/3 & 3/3 \\
Added & Value / unit & 3 & 3 & 0/3 & 3/3 \\
Added & Object association & 3 & 3 & 0/3 & 3/3 \\
Added & Condition / negation & 3 & 3 & 0/3 & 3/3 \\
\textbf{Added total} & --- & 12 & 12 & 0/12 & 12/12 \\
\bottomrule
\end{tabular}
\end{table}

As shown in Table~\ref{tab:d4-error-detection}, both settings detected all
11 eligible errors involving shared facts. For added facts, the restricted
checklist detected 0 of 12 errors, while the full D4 checklist detected all
12. Across the four tested error types, broader checklist coverage enabled
detection of errors involving the added facts in these controlled pairs.

\subsection{Validation and Specification Freezing}
\label{app:cb-validation}

Structural validation checks source quotations, target identifiers, required
fields, and duplicate entries. Semantic review examines whether requirements
follow from the final input, preserve necessary bindings, and belong to the
intended component. Coverage review identifies omitted requirements, while
revision records preserve the changes and their sources. Source ambiguity and
exclusions are recorded explicitly. Structural validity, semantic support,
and fact coverage serve distinct purposes in this process.

The released specifications are versioned with the final inputs and held
fixed across generators. Chinese and English descriptions share the same
sample-specific requirements. Changes to the generation inputs require a
corresponding specification review. Generated-image observations and scores
are obtained separately under the evaluation protocol in
Appendix~\ref{app:eval-scoring}.

\section{D3 and D4 Evaluation Details}
\label{app:eval-scoring}

The frozen specifications define the presentation and semantic requirements
for each sample. Evaluation uses evidence from the generated image;
reference-image measurements are not scoring targets.

\subsection{D3: Layout and Style Evaluation}
\label{app:eval-evidence}
\label{app:eval-d3-rules}
\label{app:eval-d3-aggregation}

\paragraph{Observation extraction.}
The evaluator receives the generated image, the frozen checklist, and a target
inventory containing text identities and region membership. A VLM associates
visible text with the targets named in each check, using OCR candidates when
available. Repeated strings are distinguished through their region and
surrounding context. Target absence requires visual confirmation beyond an omitted OCR detection.
Each judgment records the observed property, supporting target boxes, and an
explanation relative to the specified criterion and quantifier.

\paragraph{Per-requirement judgments.}
Each check receives \texttt{pass}, \texttt{fail}, or \texttt{unknown}.
Table~\ref{tab:eval-d3-rules} summarizes the six attribute families.
A requirement on every target must be supported over that scope; a global
requirement concerns the specified group or region as a whole. Apart from the
supported numerical placement tests, the judgments are made by the VLM from
the supplied criterion and visible evidence. Numeric thresholds are used only
when provided by the criterion. Clear satisfaction yields a pass and a clear
contradiction yields a fail; insufficient evidence remains unknown. Unspecified
attributes do not introduce additional constraints.

\begin{table}[!htbp]
\centering\small
\setlength{\tabcolsep}{4pt}
\renewcommand{\arraystretch}{1.15}
\caption{\textbf{D3 checklist judgments.} Each check assesses the property
and target scope defined by its frozen criterion.}
\label{tab:eval-d3-rules}
\begin{tabularx}{\linewidth}{@{}p{0.15\linewidth}X@{}}
\hline
Attribute & Assessment \\
\hline
Placement & Associate the requested text or region with its visible extent.
For supported explicit canvas ranges, the program tests the verified target
center against the allowed range and tolerance. Each-target checks test all
specified centers; global checks use their union center. Other location
criteria receive a VLM judgment against the named spatial context. \\
Spatial/reading order & The VLM identifies the requested instances and compares
their sequence in the specified direction. For a left-to-right sequence
$A,B,C$, it checks whether these instances appear in that order; for table
rows or columns, it checks the requested row or column correspondence.
The explanation identifies the supporting or conflicting arrangement. \\
Size hierarchy & The VLM compares the visible character scale of the specified
subject and comparison texts. For ``title larger than body,'' evidence must
identify both title and body and describe their relative character size;
the total area of a longer text block is not a font-size measure. A multi-level
requirement checks the requested title--body--caption ordering. Qualitative
size judgments are visual comparisons, not recovered font point sizes. \\
Orientation & The VLM distinguishes the direction in which characters form a
line from the orientation of the glyphs themselves. Upright characters stacked
vertically differ from a horizontal label rotated by $90^\circ$. For a rotated
axis label, it inspects the glyphs and baseline relative to the image axes;
for horizontal text, it inspects the line direction. A tall box alone is not
rotation evidence. Numeric angles are assessed against the supplied criterion,
not by a separate angle-estimation module. \\
Foreground color & The VLM inspects the visible letter strokes and compares
them with the requested named color and modifiers such as light or dark.
Backgrounds, shadows, and surrounding graphics are distinguished from the
lettering. A group requirement covers the specified members and exceptions;
each specified member must satisfy the requested color.
The evaluator does not apply a universal RGB/HSV scoring threshold. \\
Font style & The VLM inspects visible glyph features corresponding to the
requested style, such as terminal serifs, stroke weight, slant, handwritten
forms, or hollow outlines. Each requested property is judged separately;
for example, bold lettering does not by itself establish a serif typeface.
The evidence describes the observed feature rather than substituting a general
aesthetic judgment. \\
\hline
\end{tabularx}
\end{table}

For supported numerical placement checks, boxes use normalized
$[x_1,y_1,x_2,y_2]$ coordinates. Each-target criteria test individual target
centers; global criteria test the center of the union of the specified targets.
Region boxes must cover the required region. Centers are compared with the
allowed range and tolerance stored in the specification. Numerical placement judgments require verified target geometry.

\paragraph{Verification and reinspection.}
Under full verification, a second independent assessment examines every check
without seeing the preceding answer. Conflicting passing and failing judgments
receive a third assessment; two agreeing evidence-valid votes determine the
result. Unresolved checks undergo bounded reinspection using full-image context
and local detail crops where a valid candidate region is available. Coordinates
remain relative to the full image. Malformed coordinates are requested again
rather than converted by assuming an unspecified coordinate convention.
Failed batches may be split into smaller groups, and technical failures receive
bounded retries. The verification and retry settings accompany the results.

Confirmed missing targets fail the corresponding requirement. Confirmed
generation illegibility can justify failure when it prevents satisfaction of
that criterion; it does not invalidate an otherwise observable color, direction,
or placement. Uncertain correspondence, insufficient evidence, and technical
errors remain unknown.

\paragraph{D3 aggregation.}
Let $\mathcal C_{ik}$ denote the checks in attribute family $k$ for sample $i$.
Once all applicable checks are resolved, their binary judgments
$s_{ic}\in\{0,1\}$ give
\begin{equation}
 a_{ik}=\frac{\sum_{c\in\mathcal C_{ik}}s_{ic}}{|\mathcal C_{ik}|},
 \qquad
 D3_i=\frac{100}{|\mathcal A_i|}\sum_{k\in\mathcal A_i}a_{ik},
\end{equation}
where $\mathcal A_i$ contains the families with at least one check.
The attribute score describes the proportion of satisfied requirements.
For example, $a_{ik}=0.5$ means that half of the checks in that attribute
family pass and half fail; it is not a partial judgment on an individual
check. Values between $0$ and $1$ follow from the number of satisfied checks.
Active families receive equal weight, and families without checks are excluded.

\subsection{D4: Semantic Evaluation}
\label{app:eval-other-scores}

\paragraph{Evaluation inputs and outputs.}
The evaluator receives the generated image and the frozen D4 checklist,
including the required meanings, targets, and satisfaction criteria. The five
components are core textual semantics, communicative intent, text--image
consistency, semantic logic, and physical-carrier integration. Each applicable
check produces an evidence record and a judgment; these are aggregated into
component scores and a D4 total when all checks are resolved.

\paragraph{Step 1: evidence extraction.}
The VLM records the actual text read, visible scene content, and inspected
scope for each requirement. Core facts retain the required subjects, attributes,
values, units, conditions, negations, and associations. Text--image and logic
checks additionally record the participating entities and observed relation.
Carrier checks inspect attachment to the specified physical surface and its
perspective, curvature, material, and lighting where required. Bounding boxes
are optional for D4 judgments.

\paragraph{Step 2: evidence validation.}
Validation checks identities, verdicts, observation status, and required evidence
fields. Passing judgments require observed supporting evidence. Confirmed
omissions and incorrect facts fail; confirmed generation illegibility also
fails when the required meaning cannot be recovered. Carrier integration is assessed from the visible attachment and surface
relation. Incomplete or contradictory
evidence remains unknown. Under full verification, two independent assessments
examine every check, and disagreements or unknown answers receive a third
assessment. Two agreeing evidence-valid votes determine the verdict. Remaining
unknowns receive bounded single-check reinspection and independent verification
when enabled. Technical failures are retried without being interpreted as
generation defects.

\paragraph{Step 3: component scoring and aggregation.}
Each resolved check contributes $q_{ic}=1$ if satisfied and $q_{ic}=0$
otherwise. For component $k$ with check set $\mathcal F_{ik}$, the satisfaction
proportion and D4 score are
\begin{equation}
 r_{ik}=\frac{\sum_{c\in\mathcal F_{ik}}q_{ic}}{|\mathcal F_{ik}|},
 \qquad
 D4_i=100\sum_{k\in\mathcal B_i}\widetilde v_{ik}r_{ik},
 \qquad
 \widetilde v_{ik}=\frac{v_k}{\sum_{j\in\mathcal B_i}v_j}.
\end{equation}
The weights $v=(0.45,0.10,0.20,0.20,0.05)$ correspond to core textual
semantics, communicative intent, text--image consistency, semantic logic,
and physical-carrier integration. A component score of $0.5$ indicates that
half of its requirements are satisfied; intermediate scores arise from
averaging check-level judgments. Components without applicable checks are
excluded and the remaining weights are renormalized. Unknown judgments are
not assigned $0.5$ and do not remove checks or components from the denominator.
Supporting observations and check-level judgments accompany the scores.

\paragraph{Unresolved checks and reporting.}
\label{app:eval-unresolved}
For an active attribute family or semantic component, let $P_{ik}$, $F_{ik}$,
and $U_{ik}$ denote passing, failing, and unknown checks, with
$N_{ik}=P_{ik}+F_{ik}+U_{ik}$. Check coverage and score bounds are
\begin{equation}
 C_{ik}=\frac{P_{ik}+F_{ik}}{N_{ik}},\qquad
 L_{ik}=\frac{P_{ik}}{N_{ik}},\qquad
 H_{ik}=\frac{P_{ik}+U_{ik}}{N_{ik}}.
\end{equation}
Passing and failing checks both count as resolved. For D3, image-level
bounds are
\begin{equation}
 L_i^{(3)}=\frac{100}{|\mathcal A_i|}\sum_{k\in\mathcal A_i}L_{ik},
 \qquad
 H_i^{(3)}=\frac{100}{|\mathcal A_i|}\sum_{k\in\mathcal A_i}H_{ik}.
\end{equation}
For D4, they are
\begin{equation}
 L_i^{(4)}=100\sum_{k\in\mathcal B_i}\widetilde v_{ik}L_{ik},
 \qquad
 H_i^{(4)}=100\sum_{k\in\mathcal B_i}\widetilde v_{ik}H_{ik}.
\end{equation}
These bounds assign unknowns their minimum or maximum contribution; they are
not statistical confidence intervals. Image-level coverage follows the same
attribute or component weighting, and model-level coverage averages images.
Complete-score means and bounds over all applicable evaluated images use
separate image pools. Technical failure rates count evaluated images retaining
an unresolved request or response-format failure after retries, separately
from visual uncertainty. Appendix~\ref{app:evaluation-coverage} reports the
resulting coverage and sensitivity of model rankings.

\begingroup
\section{Additional Experiments}
\label{app:additional-experiments}
\subsection{Human Evaluation Protocol and Reliability}
\label{app:human-evaluation}
\paragraph{(1) Sample selection and evaluation scale.}
We used a two-stage sampling procedure. First, we drew 200 prompts through proportional stratified sampling by domain and subcategory from samples with complete generation results and matching generation-prompt records in the initial five-model pool. We then selected 100 prompts covering all eight domains and 38 subcategories, with 50 prompts per language. Within this coverage constraint, we prioritized samples with more required text. The resulting sample is category-covering and language-balanced, with an emphasis on long text. The same selected prompt--language pairs were evaluated across nine generators, yielding 900 images. Each image received three independent ratings on D1--D4, giving 2,700 valid rating records and 10,800 dimension scores.

\paragraph{(2) Annotators and assessment procedure.}
Six graduate students able to read both Chinese and English participated. Before annotation, we held a joint briefing to explain the four dimensions and their scoring criteria. Annotators scored independently without discussing their judgments. Model identities, automatic scores, and other annotators' ratings were hidden, and model outputs were presented in randomized order. All ratings were submitted by the annotators themselves; the organizer consolidated the records.

Annotators inspected the complete generation prompt, required-text inventory, and generated image. They assessed \textbf{D1, Glyph Structure}, for legibility and intact glyphs, including broken or merged strokes, malformed characters, and blur; \textbf{D2, Textual Correctness}, for omissions, substitutions, extra text, numbers, punctuation, capitalization, and repeated instances; \textbf{D3, Layout and Style}, for explicitly requested position, order, hierarchy, orientation, color, and font/style; and \textbf{D4, Semantic Alignment}, for required entities, attributes, quantities, relationships, communicative intent, and associations between text and depicted objects. D3 did not penalize personal aesthetic preferences or reference-image details absent from the prompt. Text errors affected D4 when they changed the required meaning or association. Annotators applied the dimension-wise rubric to each image.

Each dimension used an ordinal five-point scale: 1 indicates severe failure; 2, major problems; 3, partial fulfillment; 4, mostly fulfilled requirements; and 5, fully or almost fully fulfilled requirements. Figure~\ref{fig:human-eval-interface} illustrates the rating interface using an English long-text example, with nine anonymous outputs and independent D1--D4 rating options for each image.

\paragraph{(3) Score aggregation and model ranking.}
For each image, we average the three ratings and then take an equal-weight average across D1--D4 to obtain a human composite score. Within each language, we average the image scores over the 50 prompts for each generator and rank the nine generators by their resulting human composite scores. We compare this ranking with the ranking based on automatic composite scores over the benchmark. The human ratings were collected on the selected generated-image subset; this comparison evaluates model-level rank agreement.

\paragraph{(4) Inter-annotator agreement.}
We use ordinal Krippendorff's $\alpha$ to measure agreement between annotators rating the same images on the same dimension. Table~\ref{tab:human-agreement} reports the results.

\par\medskip\noindent
\begin{minipage}{\linewidth}
\makeatletter\def\@captype{table}\makeatother
\centering
\caption{\textnormal{Inter-annotator agreement measured by ordinal Krippendorff's $\alpha$. Agreement is computed across annotators for each dimension.}}
\label{tab:human-agreement}
\begin{tabular}{lrrrr}
\toprule
Language & D1 & D2 & D3 & D4 \\
\midrule
English & 0.783 & 0.761 & 0.622 & 0.722 \\
Chinese & 0.736 & 0.701 & 0.683 & 0.720 \\
\bottomrule
\end{tabular}
\end{minipage}
\par\medskip

\paragraph{(5) Ranking correlation and statistical uncertainty.}
We measure agreement between human and automatic composite-score rankings
using Spearman's correlation. To estimate sampling uncertainty, we use
20,000 prompt-cluster bootstrap resamples. In each language, each resample
draws 50 prompts with replacement and retains all generators and ratings
associated with each sampled prompt. The human ranking is recomputed against
the fixed automatic ranking. The 2.5th and 97.5th percentiles define the
95\% interval in Table~\ref{tab:human-ranking-ci}. This procedure quantifies
variation due to prompt sampling, conditional on the evaluated generators
and automatic scores.

\par\medskip\noindent
\begin{minipage}{\linewidth}
\makeatletter\def\@captype{table}\makeatother
\centering
\caption{\textnormal{Agreement between human and automatic overall model rankings, with 95\% prompt-cluster bootstrap percentile intervals.}}
\label{tab:human-ranking-ci}
\begin{tabular}{lcc}
\toprule
Language & Spearman's $\rho$ & 95\% confidence interval \\
\midrule
English & 0.983 & $[0.950,0.983]$ \\
Chinese & 0.983 & $[0.950,0.983]$ \\
\bottomrule
\end{tabular}
\end{minipage}
\par\medskip

\par\medskip\noindent
\begin{minipage}{\linewidth}
\makeatletter\def\@captype{figure}\makeatother
\centering
\includegraphics[width=\linewidth]{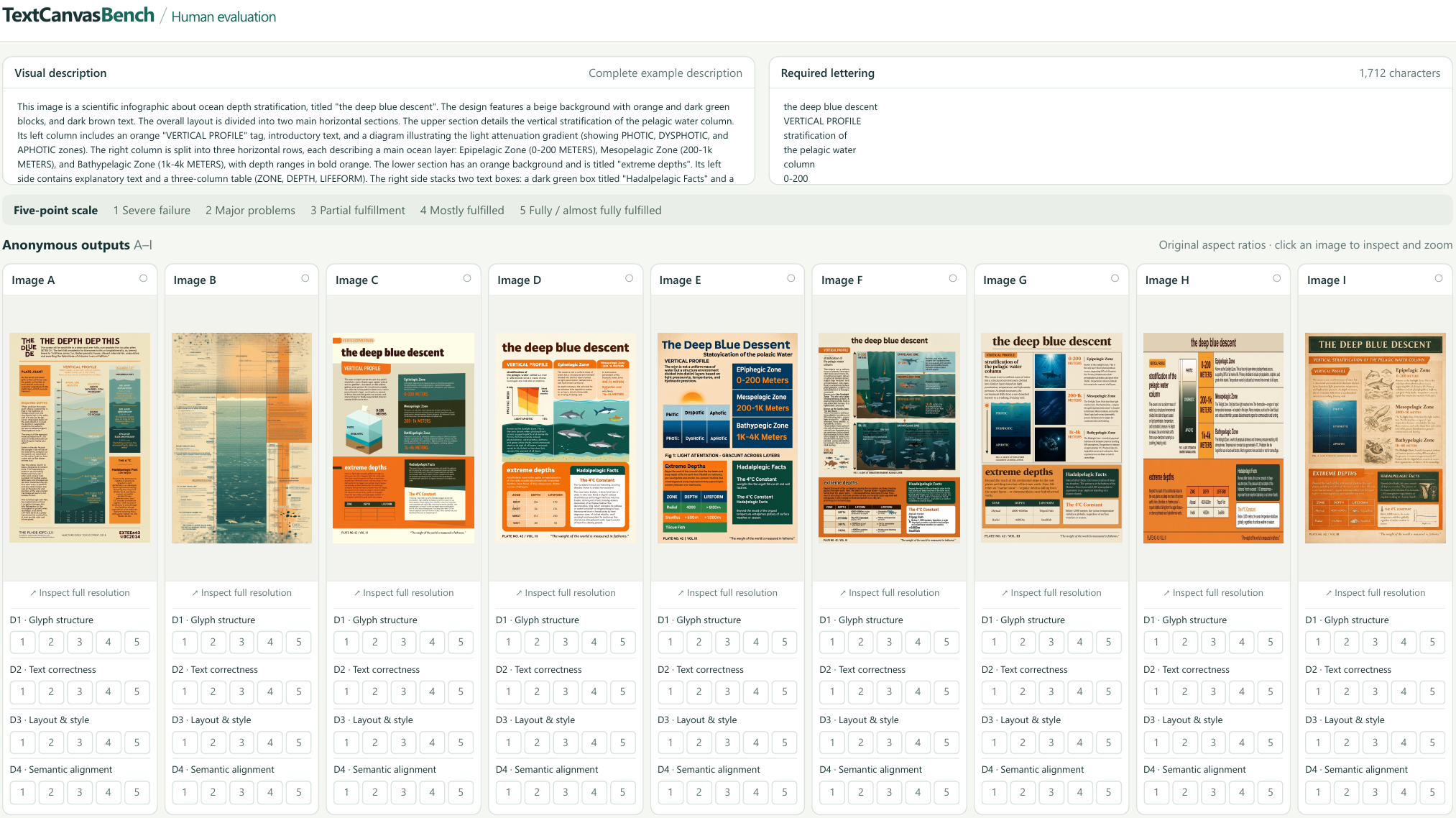}
\caption{\textnormal{Illustration of the human-evaluation interface using an English long-text example. The interface presents the visual description, required lettering, nine anonymous outputs (Images A--I), and independent five-point D1--D4 ratings. The static illustration shows input excerpts; the interface supports scrolling through complete inputs and enlarging images.}}
\label{fig:human-eval-interface}
\end{minipage}
\par\medskip
\endgroup

\subsection{Evaluation Coverage and Sensitivity to Unresolved Checks}
\label{app:evaluation-coverage}

\paragraph{(1) Coverage and technical failures.}
Confirmed omissions and requirement violations receive failing judgments;
they are resolved checks, rather than missing evaluation evidence.
Initially unresolved checks undergo targeted reinspection using contextual
target association, original-resolution inspection, and local detail views,
as described in Appendix~\ref{app:eval-d3-rules}. Technical request and
output-format failures receive bounded retries. Checks that remain
indeterminate retain an \texttt{unknown} status and remain in the checklist
denominator.

Table~\ref{tab:eval-failure-coverage} reports D3 and D4 scores, check coverage,
and residual technical failure rates. Check coverage is the resolved fraction
of applicable requirements, aggregated with the same attribute or component
weights used for scoring and then averaged over images. The technical failure
rate is the proportion of evaluated images retaining at least one unresolved
check due to a request or response-format failure after retries. It excludes
confirmed generation defects and visually indeterminate evidence. D3 coverage
ranges from 98.45\% to 99.38\%, and D4 coverage from 99.53\% to 99.94\%.
Residual technical failure rates are at most 0.85\% for D3 and 1.34\% for D4.

\par\medskip\noindent
\begin{minipage}{\linewidth}
\makeatletter\def\@captype{table}\makeatother
\centering
\caption{D3 and D4 evaluation results, with English and Chinese images pooled.
Scores use a 0--100 scale; coverage and technical failure rates are percentages.
Mean scores use images with complete judgments. Bounds aggregate all applicable
images, including those with unresolved checks; they therefore bound the
all-image mean, not the complete-image mean.}
\label{tab:eval-failure-coverage}
\small
\setlength{\tabcolsep}{3pt}
\resizebox{\linewidth}{!}{%
\begin{tabular}{lrrrrrrrr}
\toprule
& \multicolumn{4}{c}{D3} & \multicolumn{4}{c}{D4} \\
\cmidrule(lr){2-5}\cmidrule(lr){6-9}
Model & Mean & Coverage & Tech. failure & Bounds & Mean & Coverage & Tech. failure & Bounds \\
\midrule
GPT-Image-2 & 97.20 & 99.15 & 0.85 & 96.32--97.17 & 99.01 & 99.93 & 0.22 & 98.90--98.98 \\
Seedream-5.0-Pro & 95.68 & 99.09 & 0.58 & 94.76--95.67 & 91.69 & 99.92 & 0.39 & 91.50--91.59 \\
Nano-Banana-Pro & 94.94 & 99.18 & 0.13 & 94.15--94.97 & 97.14 & 99.94 & 0.22 & 97.01--97.07 \\
Boogu Turbo & 94.61 & 99.34 & 0.25 & 93.85--94.51 & 74.09 & 99.85 & 1.34 & 73.44--73.60 \\
Z-Image-Turbo & 91.42 & 99.38 & 0.16 & 90.68--91.30 & 80.65 & 99.87 & 0.94 & 80.09--80.21 \\
Qwen-Image & 90.34 & 99.24 & 0.16 & 89.22--89.97 & 61.54 & 99.53 & 1.09 & 60.24--60.71 \\
GLM-Image & 79.79 & 98.45 & 0.28 & 77.43--78.98 & 60.67 & 99.87 & 0.19 & 60.41--60.54 \\
FLUX.1 & 31.22 & 99.12 & 0.12 & 32.07--32.95 & 10.69 & 99.87 & 0.25 & 10.69--10.82 \\
SD3.5 & 20.63 & 98.52 & 0.25 & 21.66--23.14 & 4.35 & 99.74 & 0.16 & 4.30--4.56 \\
\bottomrule
\end{tabular}}
\end{minipage}
\par\medskip

\paragraph{(2) Sensitivity of scores and rankings.}
For each image, we hold resolved judgments fixed and assign every unresolved
check its minimum or maximum contribution. Applying the scoring weights and
averaging over the same image pool yields lower and upper model-score bounds.
These are sensitivity bounds, not statistical confidence intervals. Unlike
complete-image means, they include images with unresolved checks.

The D3 intervals have widths of 0.62--1.55 points, and the D4 intervals have
widths of 0.06--0.47 points. For D3, 34 of 36 pairwise comparisons have disjoint intervals. The two
overlapping pairs involve Nano-Banana-Pro versus Seedream-5.0-Pro and
Nano-Banana-Pro versus Boogu Turbo. The ordering of the other pairs is
preserved under all admissible assignments to unresolved checks. For D4, 35 of 36 pairwise comparisons have disjoint
intervals. Qwen-Image and GLM-Image are the sole exception: their intervals
overlap, and their relative order is not determined independently of the
unresolved checks. Thus, residual unknowns have a limited effect on aggregate
scores, while closely matched models require more cautious rank comparisons.

\paragraph{(3) Diagnostic contribution of D3 and D4.}
D3 and D4 identify errors that glyph quality and string-level correctness do
not fully describe. For example, Boogu Turbo has a higher D3 score than
Z-Image-Turbo, whereas the ordering reverses for D4, as shown in
Table~\ref{tab:eval-failure-coverage}. This distinction separates adherence
to presentation requirements from preservation of semantic associations.
The controlled edits in Figure~\ref{fig:exp-diagnostics} provide complementary
evidence: changing title color lowers the targeted D3 score by 94.8 points,
whereas swapping text--object bindings lowers the targeted D4 score by
100.0 points with only a 0.2-point change in D2. Meaning-preserving spacing
changes leave the tested D4 association score unchanged. These responses
support the distinct diagnostic roles of the two dimensions;
Appendix~\ref{app:diagnostic-design} details the controlled comparisons.

\begingroup

\subsection{Paired Core-Information Diagnostics and Model Design}
\label{app:diagnostic-design}

\paragraph{(1) Design and audit scope.}
We use all 12 scenes (six Chinese, six English) in the existing generation
experiment, with GPT-Image-2, Nano-Banana-Pro, Seedream-5.0-Pro, Qwen-Image,
and Boogu-Image Turbo. The selected conditions cross target character count
$N\in\{256,1024\}$ with requested independent text-container count
$K\in\{4,16\}$, using one existing output per condition:
$12\times5\times2\times2=240$ images. These endpoints are selected by
condition, not output score. $K$ denotes requested containers, not realized
text blocks. Increasing $N$ adds surrounding text while preserving four
predefined core facts; changing $K$ preserves the complete required text.
Text-load comparisons match scene, model, requested $K$, and actual output
resolution. The core facts were specified before this audit and bind subjects
to content, numerical values, or units. Each model contributes 48 images and
192 fact instances. Visual audit records correct content and associations,
missing content, readable substitutions, glyph damage, definite binding
errors/conflicts, and unresolved evidence. Fidelity is the proportion of
resolved instances with correct content and associations, not full-text
accuracy or the complete D4 score. The visual audit records the evidence supporting each judgment.

\paragraph{(2) What the examples show.}
In Figure~\ref{fig:additional-content-errors}b, the same Seedream menu scene
requires RMB~12 per skewer at $N=256$ and $N=1024$, with matched requested
$K$. The lower-text-load output retains the unit; the higher-text-load output
replaces it with ``per portion.'' The glyphs remain readable, but the pricing
unit changes. Boogu's whole-grain example requires Brown Rice: 37\%, yet
assigns Brown Rice both 43\% and 37\%. The requested strings appear, but the
associations conflict. Across Boogu's audited outputs, 81 instances have
missing/substituted content and 12 have definite binding errors/conflicts.
Among its 21 images with available $D1\ge90$, 15 contain a readable
substitution or definite binding error. This separates glyph quality from
field identity and relational fidelity.

\paragraph{(3) Controlled evaluator-sensitivity analysis.}
Figure~\ref{fig:exp-diagnostics}b uses a separate evaluator-sensitivity test:
24 dataset reference images each have an unedited version, a specified
heading-color edit, a text-region association swap, and a valid spacing
redesign (96 images). The four versions share the same requirements.
Association swaps preserve strings while changing subject--value/label
relations. Their mean D2 change is $-0.2$ points, whereas the targeted D4
association score falls by 100.0 points. Color edits lower the targeted D3
color score by 94.8 points and leave the D4 association score unchanged;
valid redesigns also leave the latter unchanged. D3 checks a specified text
color and D4 checks two semantic associations, each on a 0--100 scale. Each D2/D3 edit condition has 24 valid pairs;
D4 color, association, and redesign conditions have 23, 21, and 23 pairs,
respectively. Unresolved pairs are excluded. These controls distinguish evaluator responses to strings, presentation,
and semantic associations.

\paragraph{(4) Fixed core facts under increased text load.}
Figure~\ref{fig:additional-paired-effects}a compares $N=256$ and $N=1024$.
Fidelity decreases by 12.6 points for Seedream and 22.9 for Boogu, with
exploratory 95\% scene-bootstrap intervals $[-24.0,-3.4]$ and
$[-40.8,-4.4]$. GPT-Image-2 and Nano-Banana-Pro change by $-3.1$ and $+1.0$
points, retaining 189 and 191 of their 192 audited core-fact instances.
Paired changes use facts resolved at both endpoints, average within scene
and then across scenes, and use 20,000 scene bootstrap resamples. Each
condition contains one existing output. Seedream's and Boogu's decreases are
23.2 and 45.3 points in the Chinese scene group, versus 2.1 and 0.6 in the
English group. Figure~\ref{fig:additional-paired-effects}b reports the
additional fixed-text container-count contrast: Seedream changes by $+6.4$
points ($[-0.9,+15.8]$), and Boogu by $-14.3$ ($[-30.8,+2.8]$). The effects of increasing $K$ differ in direction, and both intervals
include zero. Qwen remains near the score floor in both conditions.

\par\medskip\noindent
\begin{minipage}{\linewidth}
\makeatletter\def\@captype{figure}\makeatother
\centering
\includegraphics[width=\linewidth]{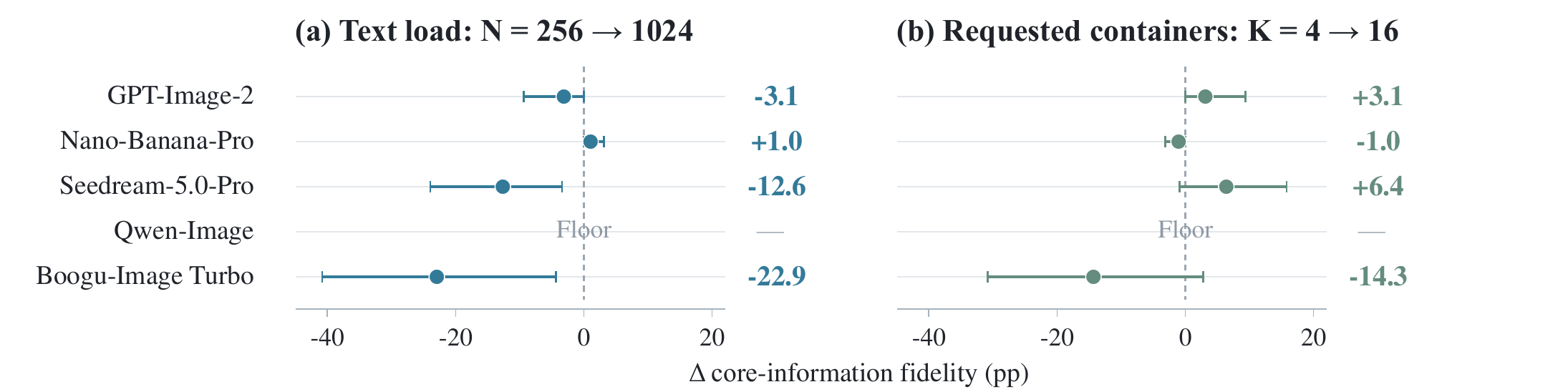}
\caption{Paired changes in fixed core-information fidelity. Panel a increases target text from $N=256$ to $N=1024$ at matched requested $K$ and actual resolution; panel b increases requested $K$ from 4 to 16 with identical required text. Points are scene-averaged high-minus-low differences; intervals are exploratory 95\% scene-bootstrap intervals. Facts must be resolved at both endpoints. Each condition uses one existing output.}
\label{fig:additional-paired-effects}
\end{minipage}
\par\medskip

\paragraph{(5) Design implications.}
Table~\ref{tab:additional-model-design} relates public designs to the two
observed retention problems.

\textbf{Architecture and text representation.}
Glyph encoding, joint text--visual processing, and character or paragraph
coverage address complementary aspects of generation. Readable substitutions
and conflicting Brown Rice associations show that local rendering and string
presence can coexist with incorrect field identity. The design objective
is therefore to preserve text-instance identities and subject--value--unit
relations alongside glyph features throughout visual generation. Repeated labels
and similar numerical values require distinct instance identities.

\textbf{Training data and supervision.}
The fixed-fact comparison supplies a paired data-construction condition:
expand surrounding required text while preserving the same subjects, values,
and units. Supervision can check unchanged-field retention as well as the
accuracy of added text, including omissions, unit substitutions, and conflicts.
This supplements character coverage and paragraph/template tasks with an
explicit content-preservation objective, motivated by the observed
retention errors.

\textbf{Planning, rendering, and correction.}
Planning can enumerate fields and relations; layout assignment can retain
their identities; rendering and subsequent corrections can recheck units,
values, duplicates, and associations against the same inventory. The testable
objective is that adding content should preserve existing core facts.
These directions draw on public model designs and observed output errors;
specific module and training-data effects require controlled ablations.

\par\medskip\noindent
\begin{minipage}{\linewidth}
\makeatletter\def\@captype{table}\makeatother
\centering
\caption{Public model designs, output evidence, and implications for content preservation.}
\label{tab:additional-model-design}
\begingroup
\fontsize{8.5}{10.2}\selectfont
\setlength{\tabcolsep}{4pt}
\renewcommand{\arraystretch}{1.12}
\begin{tabularx}{\linewidth}{@{}>{\raggedright\arraybackslash}p{0.20\linewidth}>{\raggedright\arraybackslash}p{0.355\linewidth}>{\raggedright\arraybackslash}X@{}}
\toprule
\textbf{Model} & \textbf{Public design / output evidence} & \textbf{Design implication} \\
\midrule
Qwen-Image\newline {\scriptsize\citep{wu2025qwenimage}}
& Qwen2.5-VL + dual-stream MMDiT; paragraph / template curriculum.
& Preserve exact coverage of required text instances during conditioning. \\
\addlinespace[3pt]
Z-Image-Turbo\newline {\scriptsize\citep{tongyi2025zimage}}
& Single-stream S3-DiT; Turbo uses 8 function evaluations.
& Couple joint token processing with full-text identity and coverage. \\
\addlinespace[3pt]
GLM-Image\newline {\scriptsize\citep{zai2026glmimage}}
& Autoregressive generator + diffusion decoder; dedicated Glyph Encoder.
& Combine glyph rendering with global content and relation constraints. \\
\addlinespace[3pt]
Boogu-Image Turbo\newline {\scriptsize\citep{chen2026booguimage}}
& Instruction encoder + DiT; targeted character coverage.
& Extend character coverage to entity--value--unit consistency. \\
\midrule
GPT-Image-2
& 189/192 correct core anchors; text-load change: $-3.1$~pp.
& Maintain required instance identity under increased text load. \\
\addlinespace[3pt]
Nano-Banana-Pro
& 191/192 correct core anchors; text-load change: $+1.0$~pp.
& Retain core content across text and organization requirements. \\
\addlinespace[3pt]
Seedream-5.0-Pro
& 8 readable substitutions; text-load change: $-12.6$~pp.
& Protect value and unit identity as surrounding text expands. \\
\bottomrule
\end{tabularx}
\par\vspace{3pt}
\begin{minipage}{\linewidth}
\fontsize{7.5}{9.0}\selectfont
Output evidence is from the 240-image diagnostic subset. Text-load change is
the scene-averaged paired change in core-information fidelity as target
character count increases from 256 to 1024, at fixed requested container count
and actual output resolution; pp denotes percentage points.
\end{minipage}
\endgroup
\end{minipage}
\par\medskip
\endgroup

\subsection{Stability across Repeated Generations}
\label{app:ranking-robustness}

\paragraph{Evaluation setting.}
We assess generation variability using three independent generations for each
of the same 200 TextCanvasBench prompts and nine generators used in
Figure~\ref{fig:exp-benchmark-ablation}. The subset emphasizes longer required
text. Model versions, generation parameters, prompts, and evaluation settings
are held fixed across runs. All outputs are evaluated with the same full-image
TextPecker D1 protocol. Table~\ref{tab:robustness-d1-runs} reports the model
means for each run; Figure~\ref{fig:exp-benchmark-ablation} uses their average.

\par\medskip\noindent
\begin{minipage}{\linewidth}
\makeatletter\def\@captype{table}\makeatother
\centering
\caption{D1 stability across three independent generations of the same 200
prompts. Mean and range summarize the three run-level means on a 0--100 scale.
Z-Image and Z-Image-Turbo are distinct generators in this experiment.}
\label{tab:robustness-d1-runs}
\small
\setlength{\tabcolsep}{6pt}
\begin{tabular}{lrrrrr}
\toprule
Model & Run 1 & Run 2 & Run 3 & Mean & Range \\
\midrule
GPT-Image-2 & 98.29 & 97.97 & 97.50 & 97.92 & 0.79 \\
Nano-Banana-Pro & 98.33 & 97.68 & 96.18 & 97.40 & 2.15 \\
Seedream-5.0-Pro & 92.63 & 92.78 & 94.62 & 93.34 & 1.99 \\
GLM-Image & 92.46 & 92.51 & 92.97 & 92.65 & 0.51 \\
Z-Image & 90.78 & 90.46 & 90.22 & 90.49 & 0.56 \\
Z-Image-Turbo & 89.69 & 89.45 & 89.34 & 89.49 & 0.35 \\
Boogu Base & 82.85 & 81.75 & 78.48 & 81.03 & 4.37 \\
Boogu Turbo & 81.66 & 80.81 & 77.89 & 80.12 & 3.77 \\
Qwen-Image & 79.14 & 80.27 & 78.07 & 79.16 & 2.20 \\
\bottomrule
\end{tabular}
\end{minipage}
\par\medskip

\paragraph{Score variability and model ordering.}
The range of a model's three run-level means is 0.35--4.37 points.
GLM-Image, Z-Image, and Z-Image-Turbo have ranges of 0.51, 0.56, and
0.35 points, respectively. Cross-model score ranges remain 17.70--19.61
points across runs. Rank changes are confined to GPT-Image-2 versus
Nano-Banana-Pro and Boogu Turbo versus Qwen-Image; the remaining five models
retain their positions. The three-run mean has the same ordering as Run 2.

Within-family comparisons retain their direction in every run. Z-Image
exceeds Z-Image-Turbo by 1.09, 1.01, and 0.88 points, while Boogu Base
exceeds Boogu Turbo by 1.19, 0.94, and 0.59 points. These repetitions show
consistent broad separation, while also revealing that close model pairs
can exchange rank across independent generations.

\end{document}